\documentclass[10pt,twocolumn,letterpaper]{article}

\usepackage{iccv}
\usepackage{times}
\usepackage{epsfig}
\usepackage{amsmath}
\usepackage{amssymb}
\usepackage{lipsum}
\usepackage{algorithm}
\usepackage{algpseudocode}
\mathchardef\mhyphen="2D

\usepackage[breaklinks=true,bookmarks=false]{hyperref}

\iccvfinalcopy 

\def\iccvPaperID{****} 

\ificcvfinal\fi

\usepackage{siunitx}    
\usepackage{array,multirow}
\usepackage{colortbl}
\usepackage{xcolor}
\usepackage{arydshln}
\usepackage{caption}
\usepackage[export]{adjustbox}
\usepackage{subcaption}

\usepackage{acro}
\DeclareAcronym{uq}{
    short=UQ,
    long=Uncertainty Quantification
}
\DeclareAcronym{bnn}{
    short=BNN,
    long=Bayesian Neural Networks
}
\DeclareAcronym{mcdropout}{
    short=MC-Dropout,
    long=Monte Carlo Dropout
}
\DeclareAcronym{vae}{
    short=VAE,
    long=Variational Autoencoders
}
\DeclareAcronym{de}{
    short=DE,
    long=Deep Ensembles
}
\DeclareAcronym{ud}{
    short=UD,
    long=Uncertainty Distillation
}
\DeclareAcronym{kd}{
    short=KD,
    long=Knowledge Distillation
}
\DeclareAcronym{dudes}{
    short=DUDES,
    long='\textbf{D}eep \textbf{U}ncertainty \textbf{D}istillation using \textbf{E}nsembles for \textbf{S}egmentation'
}
\DeclareAcronym{sgd}{
    short=SGD,
    long=Stochastic Gradient Descent
}
\DeclareAcronym{dnn}{
    short=DNN,
    long=Deep Neural Network
}
\DeclareAcronym{rmsle}{
    short=RMSLE,
    long=root mean squared logarithmic error,
}

\begin{document}

\title{U-CE: Uncertainty-aware Cross-Entropy for Semantic Segmentation}

\author{Steven Landgraf \hspace{2mm} Markus Hillemann \hspace{2mm} Kira Wursthorn \hspace{2mm} Markus Ulrich\\
Karlsruhe Institute of Technology (KIT)\\
{\tt\small (steven.landgraf, markus.hillemann, kira.wursthorn, markus.ulrich)@kit.edu}
}

\maketitle
\ificcvfinal\thispagestyle{empty}\fi

\begin{abstract}
Deep neural networks have shown exceptional performance in various tasks, but their lack of robustness, reliability, and tendency to be overconfident pose challenges for their deployment in safety-critical applications like autonomous driving. In this regard, quantifying the uncertainty inherent to a model's prediction is a promising endeavour to address these shortcomings. In this work, we present a novel \textbf{U}ncertainty-aware \textbf{C}ross-\textbf{E}ntropy loss (U-CE) that incorporates dynamic predictive uncertainties into the training process by pixel-wise weighting of the well-known cross-entropy loss (CE). Through extensive experimentation, we demonstrate the superiority of U-CE over regular CE training on two benchmark datasets, Cityscapes and ACDC, using two common backbone architectures, ResNet-18 and ResNet-101. With U-CE, we manage to train models that not only improve their segmentation performance but also provide meaningful uncertainties after training. Consequently, we contribute to the development of more robust and reliable segmentation models, ultimately advancing the state-of-the-art in safety-critical applications and beyond.
\footnote{Code will be made available once the publication process is complete.}
\end{abstract}

\section{Introduction}
Humans often make poor decisions and reach erroneous conclusions while overestimating their abilities, a phenomenon known as the Dunning-Kruger effect \cite{kruger1999unskilled}. Although deep neural networks are highly effective at solving semantic segmentation problems \cite{minaee2020ImageSegmentation}, they also suffer from overconfidence \cite{guo2017CalibrationModerna}. Additionally, neural networks lack interpretability \cite{gawlikowski2022SurveyUncertainty} and struggle to distinguish between in-domain and out-of-domain samples \cite{lee2018TrainingConfidencecalibrated}. These flaws are particularly relevant in safety-critical applications, such as autonomous driving \cite{mcallister2017ConcreteProblems} and medical imaging \cite{leibig2017LeveragingUncertainty}, as well as in computer vision tasks that have high demands on reliability, like industrial inspection \cite{steger2018MachineVision, heizmann2022implementing} and automation \cite{landgraf2023segmentation, ulrich_2021}, where robust predictions are crucial. Misclassifying pixels in these contexts can lead to severe consequences, emphasizing the need for robust and trustworthy segmentation models.

\begin{figure}[t!]
\centering
\includegraphics[width=1\linewidth]{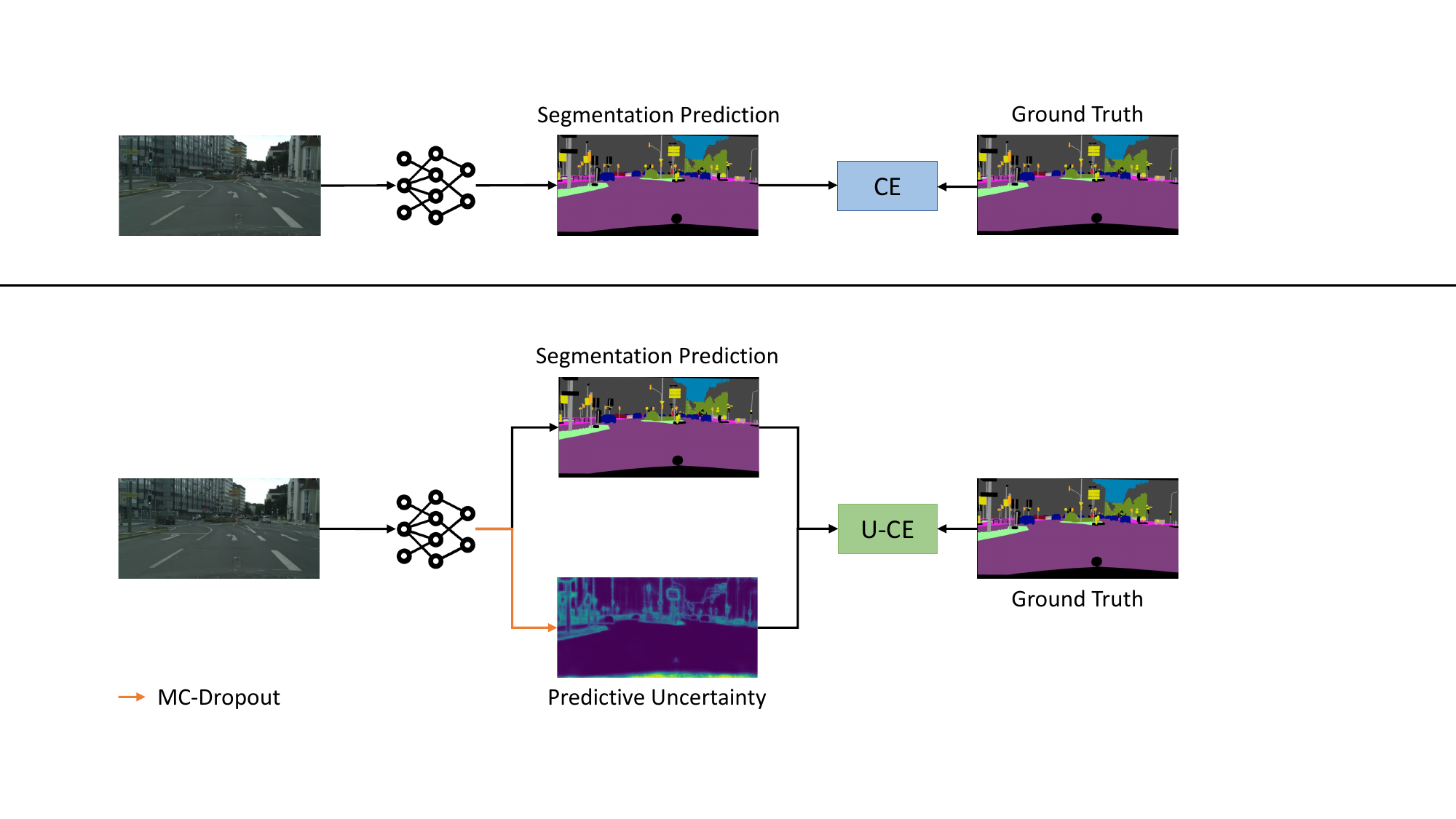}
\caption{U-CE introduces an uncertainty-aware cross-entropy loss that dynamically incorporates the predictive uncertainties provided by Monte Carlo Dropout (MC-Dropout) into the training process. As a result, we manage to train models that are naturally capable of predicting meaningful uncertainties after training while also improving their segmentation performance.}
\label{fig: 1}
\end{figure}

Previous work suggests that quantifying the uncertainty inherent to a model's prediction is a promising endeavour to enhance the safety and reliability of such applications \cite{landgraf2023dudes, leibig2017LeveragingUncertainty, lee2018TrainingConfidencecalibrated, mukhoti2018evaluating, mukhoti2023deep}. These uncertainties provide additional insights beyond the common softmax probabilities, revealing regions where the model is indecisive and likely to make errors. Surprisingly, the utilization of these uncertainties during the training of segmentation models has not been thoroughly explored.

In this work, we present a novel \textbf{U}ncertainty-aware \textbf{C}ross-\textbf{E}ntropy loss, referred to as \textbf{U-CE}, that addresses this gap by incorporating dynamic uncertainty estimates into the training process as shown in Figure \ref{fig: 1}. Through pixel-wise uncertainty weighting of the well-known cross-entropy loss (CE), we harness the valuable insights provided by the uncertainties for more effective training. With U-CE, we manage to train models that are naturally capable of predicting meaningful uncertainties after training while simultaneously improving their segmentation performance.

Our contributions can be summarized as follows: Firstly, we propose the U-CE loss function, which utilizes uncertainty estimates to guide the optimization process, emphasizing regions with high uncertainties. Secondly, we conduct extensive experiments on two benchmark datasets, Cityscapes \cite{cordts2016CityscapesDataset} and ACDC \cite{SDV21}, using two common backbones, ResNet-18 and ResNet-101 \cite{he2015DeepResidual}, demonstrating the superiority of U-CE over regular CE training. Lastly, we present additional insights, limitations, and potential improvements for U-CE through multiple ablation studies and a thorough discussion.


\section{Related Work}
In this section, we briefly review the related work on uncertainty quantification and uncertainty-aware segmentation.
\subsection{Uncertainty Quantification}
Deep neural networks, with their millions of model parameters and non-linearities, have proven effective in solving complex tasks in natural language processing \cite{otter2020survey} and computer vision, like semantic segmentation \cite{minaee2020ImageSegmentation}. Unfortunately, due to their complexity, the computation of the exact posterior probability distribution of the network's output is infeasible \cite{blundell2015BayesByBackprop,loquercio2020UncertaintyFrameworkDronet}. Consequently, approximate uncertainty quantification methods are employed to offer a practical solution to tackle the intractability of the exact posterior distribution. The most prominent methods include Bayesian Neural Networks \cite{mackay1992PracticalBayesian}, Monte Carlo Dropout \cite{gal2016DropoutBayesian}, and Deep Ensembles \cite{lakshminarayanan2017SimpleScalable}. We will refer to these methods as traditional uncertainty quantification techniques throughout the following.

A mathematically grounded, though computationally complex, approach to uncertainty quantification is provided by Bayesian Neural Networks, which transform a deterministic network into a stochastic one using probabilistic distributions placed over the activations or the weights \cite{jospin2020HandsOnBaysianNetworks}. For instance, Bayes by Backprob \cite{blundell2015BayesByBackprop} employs variational inference to learn approximate distributions over the weights. These can be used to create an ensemble of models with differently sampled weights to approximate the posterior distribution of the predictions.

Gal and Ghahramani simplify this approximation process by using Monte Carlo Dropout \cite{gal2016DropoutBayesian}. While dropout is usually applied as a regularization technique \cite{srivastava2014Dropout}, Monte Carlo Dropout uses this concept to sample from the posterior distribution of a network's prediction at test time. In its original form, Monte Carlo Dropout only captures the epistemic uncertainty inherent to the model. To obtain a more comprehensive measure of uncertainty that includes the aleatoric uncertainty, which captures the noise inherent in the observations, Monte Carlo Dropout can be combined with learned uncertainty predictions and assumed density filtering \cite{gast2018ADF, kendall2017CVUncertainties, loquercio2020UncertaintyFrameworkDronet}. 

The current state-of-the-art uncertainty quantification method are Deep Ensembles, which consist of an ensemble of trained models that generate diverse predictions at test time \cite{lakshminarayanan2017SimpleScalable}. Due to the introduction of randomness through random weight initialization or different data augmentations across ensemble members \cite{fort2020DeepEnsembles}, Deep Ensembles are well-calibrated \cite{lakshminarayanan2017SimpleScalable}. Multiple studies demonstrated that Deep Ensembles generally outperform other uncertainty quantification methods across varying tasks \cite{ovadia2019DatasetShift, wursthorn2022, gustafsson2020evaluating}. However, this performance gain is associated with high computational cost. 

In addition to the aforementioned approximate uncertainty quantification methods, there has been a growing interest in deterministic single forward-pass approaches, which offer advantages in terms of memory usage and inference time. For example, van Amersfoort \etal \cite{van2020uncertainty} and Liu \etal \cite{liu2020simple} explore the concept of distance-aware output layers. While these methods demonstrate good performance, they are not competitive with the current state-of-the-art and require significant modifications to the training process \cite{mukhoti2023deep}. Another approach, proposed by Mukhoti \etal \cite{mukhoti2023deep}, simplifies the two previous methods by employing Gaussian Discriminant Analysis for feature-space density estimation after training. Although they perform on par with Deep Ensembles in some settings, their approach still necessitates a more sophisticated training approach. Additionally, fitting the feature-space density estimator is only possible after training, which is not suitable for U-CE where meaningful uncertainties are required during training.

Overall, uncertainty quantification remains an active and evolving field of research, with various approaches offering their own advantages and disadvantages. For our specific case, Monte Carlo Dropout emerges as the preferred option due to its ease of use, minimal impact on the training process, and computational efficiency compared to Deep Ensembles. Through Monte Carlo Dropout sampling, we can compute the predictive uncertainty to apply pixel-wise weighting of the well-known cross-entropy loss. With predictive uncertainties, we refer to the standard deviation of the softmax probabilities of the predicted class provided by Monte Carlo Dropout sampling.

\begin{figure*}[h!]
\centering
\includegraphics[width=\textwidth]{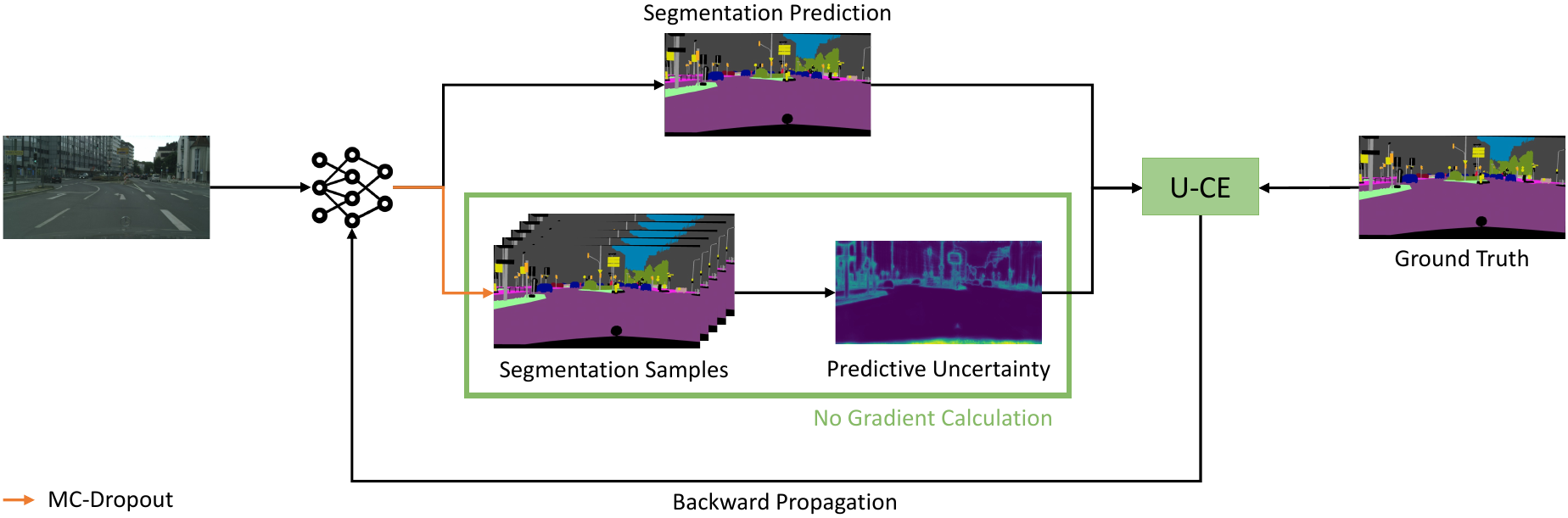}
\caption{A schematic overview of the training process of U-CE. U-CE integrates the predictive uncertainties of a Monte Carlo Dropout (MC-Dropout) model into the training process to enhance segmentation performance. In comparison to most applications of Monte Carlo Dropout, U-CE utilizes the uncertainties not only at test time but also dynamically during training by applying pixel-wise weighting to the regular cross-entropy loss.}
\label{fig: 2}
\end{figure*}

\subsection{Uncertainty-aware Segmentation}
In the domain of uncertainty-aware segmentation, researchers have explored various techniques to incorporate uncertainty measures into the training process. Surprisingly, traditional uncertainty quantification methods have been largely overlooked or underutilized. We provide an overview of notable works that leverage uncertainty-aware techniques for segmentation tasks in various domains. Additionally, we discuss how U-CE addresses the gap towards full utilization of traditional uncertainty quantification methods during training. 

Some of the earlier work on more effective training has originally been designed for object detection. For example, Lin \etal \cite{lin2017focal} introduced the focal loss that down-weights the contribution of easy examples to shift the focus more towards hard examples. Another closely related technique is online hard example mining by Shrivastava \etal \cite{shrivastava2016training}. They propose to automatically select hard examples to only learn from them and completely ignore the easy examples. By now, both methods have been successfully adapted for semantic segmentation \cite{jadon2020survey, Wang_2022_CVPR}.

Another line of work focuses on the identification and compensation of ambiguities and label noise. Kaiser \etal \cite{kaiser2023compensation} propose adding a learned bias to a network's logits and introducing a novel uncertainty branch to induce the compensation bias only to relevant regions. However, unlike U-CE, their approach does not utilize uncertainties to make training more robust, rather they aim to avoid new noise during data annotation.

More closely related to our work, Bischke \etal \cite{bischke2018segmentation} and Bressan \etal \cite{bressan2022semantic} propose to leverage uncertainties to improve training on imbalanced aerial image datasets. The former use the per-class uncertainty of the model together with the median frequency to balance training \cite{bischke2018segmentation}. We argue that dynamically weighting each pixel individually during training, which is what U-CE does, is even more valuable. The latter utilize pixel-wise weights, but only consider the class and labeling uncertainty \cite{bressan2022semantic} instead of the predictive uncertainties like U-CE.

In addition to these methods, Chen \etal \cite{chen2022hyperbolic} propose to transform the embeddings of the last layer from Euclidean space into Hyperbolic space to dynamically weight pixels based on the hyperbolic distance, which they interpret as uncertainty. Similarly, Bian \etal \cite{bian2020uncertainty} propose an uncertainty estimation and segmentation module to estimate uncertainties that they use to improve the segmentation performance. Unlike U-CE, however, these two works do not incorporate traditional uncertainty quantification methods into training.

In contrast to existing literature, U-CE fully utilizes predictive uncertainties dynamically during training. By pixel-wise uncertainty weighting of the cross-entropy loss, U-CE harnesses valuable insights from the uncertainties to guide the optimization process. This approach enables more effective training, resulting in models that are naturally capable of predicting meaningful uncertainties after training while also improving their segmentation performance.

\section{Methodology}
In the following, we provide an overview of U-CE, explain our novel uncertainty-aware cross-entropy loss and outline the implementation details.

\subsection{Overview}
The central idea of U-CE is to incorporate predictive uncertainties into the training process to enhance segmentation performance. As depicted in Figure \ref{fig: 2}, we propose two simple yet highly effective adaptions to the regular training process:
\begin{enumerate}
    \itemsep0em
    \item During training, we sample from the posterior distribution with Monte Carlo Dropout to obtain predictive uncertainties alongside the regular segmentation prediction.
    \item We apply pixel-wise weighting to the regular cross-entropy loss based on the collected uncertainties.
\end{enumerate}
To compute predictive uncertainties during training, we choose Monte Carlo Dropout. It is straightforward to implement, requires minimal tuning, and is computationally more efficient than Deep Ensembles. However, it is worth noting that other uncertainty quantification methods could also be utilized for U-CE. Exploring these alternatives is an interesting avenue for future work, which we will discuss in Section \ref{sec: discussion}.

\subsection{Uncertainty-aware Cross-Entropy} 
\begin{algorithm}[t!]
    \caption{PyTorch-like Pseudocode for a single training step with U-CE}
    \label{algo}
    \begin{algorithmic}
        \State $x, y = batch$
        \State $\hat{y} = model(x)$ 
        \State $\hat{y}_{\beta} = torch.empty(size=[\beta, \hat{y}.size()])$
            \For {$\beta$ and \textbf{with} no\_grad}
            \State $\hat{y}_{\beta}[\beta] = model(x)$
            \EndFor
        \State $p = torch.mean(softmax(\hat{y}_{\beta}), dim=0)$
        \State $q = torch.std(softmax(\hat{y}_{\beta}), dim=0)$
        \State $\sigma = q[torch.argmax(p, dim=1)]$
        \State $loss = (1 + \sigma)^\alpha \cdot torch.ce(\hat{y}, y, reduction="none")$
        \State \textbf{return} $torch.sum(loss) / torch.numel(loss)$
    \end{algorithmic}
\end{algorithm}

\textbf{Segmentation Sampling.} In contrast to typical usage of Monte Carlo Dropout, U-CE incorporates the sampling process from the posterior distribution not only at test time but also during training. To compute the necessary uncertainties for our uncertainty-aware cross-entropy loss, we perform $\beta$ sampling iterations at each training step. This generates $\beta$ segmentation samples in addition to the regular segmentation prediction. Notably, gradient computation is disabled during the sampling process as it is unnecessary for backward propagation, which relies solely on the regular segmentation prediction. By disabling gradient computation during sampling, we reduce the additional computational overhead of U-CE in terms of training time and GPU memory usage.

\textbf{Uncertainty-aware Cross-Entropy Loss.} The final objective function of U-CE builds upon the well-known categorical cross-entropy loss and can be defined as:

\begin{equation}\label{eq:l_uce}
L_{u\mhyphen ce} = - \frac{1}{N} \sum_{n=1}^{N}  w_{n} \sum_{c=1}^{C} y_{n,c} \cdot \log(p_{n, c}),
\end{equation}

where $L_{u\mhyphen ce}$ is the uncertainty-aware cross-entropy loss for a single image, $N$ is the number of pixels in the image, $C$ is the number of classes, $y_{n,c}$ is the respective ground truth label, $p_{n,c}$ is the respective predicted softmax probability, and $w_{n}$ represents the pixel-wise uncertainty weight. It is worth noting that Equation \ref{eq:l_uce} simplifies to the regular cross-entropy loss by setting $w_{n}$ to one for all pixels.

\textbf{Pixel-wise Uncertainty Weight.} The pixel-wise uncertainty weight $w_{n}$ can be formulated as:

\begin{equation}\label{eq:w}
w_{n} = (1 + \sigma_{n})^{\alpha},
\end{equation}

where $\sigma_{n}$ denotes the predictive uncertainty, and $\alpha$ controls the influence of the uncertainties in an exponential manner. The predictive uncertainty $\sigma$ represents the standard deviation of the softmax probabilities of the predicted class of the segmentation samples.

\textbf{Pseudocode.} Finally, Algorithm \ref{algo} shows how to use U-CE in the training step in a simplified way. As mentioned earlier, the two key adaptations to the regular training process are sampling with Monte Carlo Dropout during training and applying pixel-wise uncertainty weighting to the regular cross-entropy loss.

\section{Experiments}\label{experiments}
In this section, we conduct an extensive range of experiments to demonstrate the value of incorporating predictive uncertainties into the training process. Firstly, we provide quantitative results comparing regular CE to U-CE under diverse settings. Secondly, we analyze qualitative examples. Lastly, we provide multiple ablation studies.

\subsection{Setup}
\textbf{Architecture.} For all of our experiments, we employ DeepLabv3+ \cite{chen2018EncoderDecoderAtrous} as the decoder and either a ResNet-18 or ResNet-101 \cite{he2015DeepResidual} as the encoder. Both backbones are commonly used for semantic segmentation \cite{minaee2020ImageSegmentation, zhang2020comparison}, making our work highly comparable and serving as an excellent baseline for future research.

\textbf{Monte Carlo Dropout.} In order to convert our architectures into Monte Carlo Dropout models, we add a dropout layer after each of the four residual block layers of the ResNets, inspired by Kendall \etal \cite{kendall2015bayesian} and Gustafsson \etal \cite{gustafsson2020evaluating}.

\textbf{Training.} For all training processes, we use a Stochastic Gradient Descent (SGD) optimizer \cite{robbins1951StochasticApproximation} with a base learning rate of 0.01, momentum of 0.9, and weight decay of 0.0001. Additionally, we multiply the learning rate of the decoder and segmentation head by ten. Finally, we employ polynomial learning rate scheduling to decay the initial learning rate during the training process, following the formula:
\begin{equation}
lr = lr_{base} \cdot (1 - \frac{iteration}{total\:iterations})^{0.9},
\end{equation}
where $lr$ is the current learning rate, and $lr_{base}$ is the initial base learning rate. In all training processes, we use a batch size of 16 and train on four NVIDIA A100 GPUs with 40 GB of memory using mixed precision \cite{micikevicius2017mixed}.

\textbf{Datasets.} All of our experiments are based on either the Cityscapes dataset \cite{cordts2016CityscapesDataset} or the ACDC dataset \cite{SDV21}. Both datasets are publicly available street scene datasets aimed at advancing the current state-of-the-art in autonomous driving. The former consists of 2975 training images, 500 validation images, and 1525 test images. The latter contains 1600 training images, 406 validation images, and 2000 test images. Although both datasets share the same 19 evaluation classes and a void class, the ACDC dataset exclusively focuses four adverse conditions: fog, nighttime, rain, and snow.

\textbf{Data Augmentations.} To prevent overfitting, we apply a common data augmentation strategy for all training procedures, regardless of the dataset or architecture used. The strategy includes the following steps:
\begin{enumerate}
    \itemsep0em
    \item Random scaling with a factor between $0.5$ and $2.0$.
    \item Random cropping with a crop size of $768\times768$ pixels.
    \item Random horizontal flipping with a flip chance of $50\%$. 
\end{enumerate}

\textbf{Evaluation.} Since both test splits are withheld for benchmarking purposes, we utilize the validation images for testing in all our experiments. Unless otherwise specified, we only report single forward pass results based on the original validation images without resizing or sampling for a fair comparison between all of the models. Also, we set the number of segmentation samples $\beta$ to ten by default.

\textbf{Metrics.} For quantitative evaluations, we primarily report the mean Intersection over Union (mIoU), also known as the Jaccard Index, to measure the segmentation performance. In addition to the mIoU, we also utilize the Expected Calibration Error (ECE) \cite{naeini2015obtaining} to evaluate the calibration as well as the mean class-wise predictive uncertainty (mUnc) to quantitatively compare the resulting uncertainties. 

\subsection{Quantitative Evaluation}

\begin{table}[t!]
\begin{center}
\begin{adjustbox}{width=\linewidth}
\setlength\extrarowheight{1mm}
\begin{tabular}{l|c|ccc|ccc}
\multirow{2}{*}{} & \multirow{2}{*}{Encoder} & \multicolumn{3}{c|}{200 Epochs}                                         & \multicolumn{3}{c}{500 Epochs}                        \\ \cdashline{3-8}
                  &                                               & CE                   & U-CE$_{\alpha=1}$    & U-CE$_{\alpha=10}$    & CE         & U-CE$_{\alpha=1}$    & U-CE$_{\alpha=10}$    \\ \hline
Dropout (0\%)     & \multicolumn{1}{c|}{RN18}                     & \textbf{70.0}                 & -                    & -                     & \textbf{72.0}       & -                     & -                    \\
Dropout (10\%)    & \multicolumn{1}{c|}{RN18}                     & 69.4                 & 69.6               & \textbf{71.6}                  & 72.3       & 72.3               & \textbf{74.2}                 \\
Dropout (20\%)    & \multicolumn{1}{c|}{RN18}                     & 69.0               & 69.5                 & \textbf{71.8}                  & 71.9       & 72.6                  & \textbf{73.5}                 \\
Dropout (30\%)    & \multicolumn{1}{c|}{RN18}                     & 68.2                 & 69.0               & \textbf{71.0}                  & 71.9       & 72.4             & \textbf{74.1}                 \\
Dropout (40\%)    & \multicolumn{1}{c|}{RN18}                     & 66.6                 & 67.7                 & \textbf{70.5}                  & 71.1       & 71.1               & \textbf{73.7}                 \\
Dropout (50\%)    & \multicolumn{1}{c|}{RN18}                     & 64.3                 & 65.3                 & \textbf{69.6}                  & 69.0       & 69.4                  & \textbf{72.6}                 \\ \hline
Dropout (0\%)     & \multicolumn{1}{c|}{RN101}                    & \textbf{74.6}                 & -                    & -                     & \textbf{76.1}       & -                     & -                    \\
Dropout (10\%)    & \multicolumn{1}{c|}{RN101}                    & 74.8                 & 75.1                 & \textbf{76.1}                  & 76.3       & 76.6                  & \textbf{77.5}                 \\
Dropout (20\%)    & \multicolumn{1}{c|}{RN101}                    & 74.6                 & 74.8                 & \textbf{76.6}                  & 76.3       & 77.0                  & \textbf{77.7}                 \\
Dropout (30\%)    & \multicolumn{1}{c|}{RN101}                    & 74.5                 & 74.7                 & \textbf{76.1}                  & 76.4       & 76.6                  & \textbf{77.5}                 \\
Dropout (40\%)    & \multicolumn{1}{c|}{RN101}                    & 74.7                 & 74.0                 & \textbf{75.8}                  & 76.1       & 76.5                  & \textbf{78.2}                 \\
Dropout (50\%)    & \multicolumn{1}{c|}{RN101}                    & 74.1                 & 73.7                 & \textbf{75.9}                  & 76.6       & 76.6                  & \textbf{77.3}                 \\
\end{tabular}
\end{adjustbox}
\end{center}
    \caption{Quantitative comparison between regular CE and U-CE on the Cityscapes dataset \cite{cordts2016CityscapesDataset} for different dropout ratios. The provided numbers represent the mIoU $\uparrow$. Best respective results are marked in \textbf{bold}.}
    \label{tab:cityscapes}
\end{table}

\begin{table}[t!]
\begin{center}
\begin{adjustbox}{width=\linewidth}
\setlength\extrarowheight{1mm}
\begin{tabular}{l|c|ccc|ccc}
\multirow{2}{*}{} & \multicolumn{1}{c|}{\multirow{2}{*}{Encoder}} & \multicolumn{3}{c|}{200 Epochs}                                         & \multicolumn{3}{c}{500 Epochs}                        \\ \cdashline{3-8}
                  &                                               & CE                   & U-CE$_{\alpha=1}$    & U-CE$_{\alpha=10}$    & CE         & U-CE$_{\alpha=1}$    & U-CE$_{\alpha=10}$    \\ \hline
Dropout (0\%)     & \multicolumn{1}{c|}{RN18}                     & \textbf{56.3}                 & -                    & -                     & \textbf{62.2}       & -                     & -                    \\
Dropout (10\%)    & \multicolumn{1}{c|}{RN18}                     & 55.5                 & 56.4                 & \textbf{60.0}                  & 62.1       & 62.8                  & \textbf{65.0}                 \\
Dropout (20\%)    & \multicolumn{1}{c|}{RN18}                     & 54.6                 & 56.1                 & \textbf{60.5}                  & 61.5       & 62.0                  & \textbf{65.0}                 \\
Dropout (30\%)    & \multicolumn{1}{c|}{RN18}                     & 52.2                 & 54.3                 & \textbf{59.2}                  & 59.6       & 61.6                  & \textbf{64.3}                 \\
Dropout (40\%)    & \multicolumn{1}{c|}{RN18}                     & 48.9                 & 50.8                 & \textbf{58.2}                  & 56.8       & 58.8                  & \textbf{63.9}                 \\
Dropout (50\%)    & \multicolumn{1}{c|}{RN18}                     & 47.7                 & 49.3                 & \textbf{56.3}                  & 53.3       & 56.0                  & \textbf{62.4}                 \\ \hline
Dropout (0\%)     & \multicolumn{1}{c|}{RN101}                    & \textbf{65.0}                 & -                    & -                     &\textbf{ 68.8}       & -                     & -                    \\
Dropout (10\%)    & \multicolumn{1}{c|}{RN101}                    & 64.5                 & 65.3                 & \textbf{67.0}                  & 68.4       & 69.3                  & \textbf{69.9}                 \\
Dropout (20\%)    & \multicolumn{1}{c|}{RN101}                    & 64.1                 & 65.0                 & \textbf{65.8}                  & 68.5       & 68.7                  & \textbf{70.2 }                \\
Dropout (30\%)    & \multicolumn{1}{c|}{RN101}                    & 62.7                 & 64.3                 & \textbf{65.3}                  & 68.4       & 68.5                  & \textbf{69.9}                 \\
Dropout (40\%)    & \multicolumn{1}{c|}{RN101}                    & 61.1                 & 63.1                 & \textbf{65.4}                  & 67.8       & 67.8                  & \textbf{70.0}                 \\
Dropout (50\%)    & \multicolumn{1}{c|}{RN101}                    & 58.0                 & 60.2                 & \textbf{63.7}                  & 66.0       & 67.4                  & \textbf{70.2}                 \\
\end{tabular}
\end{adjustbox}
\end{center}
    \caption{Quantitative comparison between regular CE and U-CE on the ACDC dataset \cite{SDV21} for different dropout ratios. The provided numbers represent the mIoU $\uparrow$. Best respective results are marked in \textbf{bold}.}
    \label{tab:acdc}
\end{table}

\begin{table}[t!]
\begin{center}
\begin{adjustbox}{width=\linewidth}
\setlength\extrarowheight{1mm}
\setlength{\tabcolsep}{7pt}
\begin{tabular}{l|c|ccc|ccc}
\multirow{2}{*}{} & \multicolumn{1}{c|}{\multirow{2}{*}{Encoder}} & \multicolumn{3}{c|}{200 Epochs}                                         & \multicolumn{3}{c}{500 Epochs}            \\ \cdashline{3-8}
                        &                                               & mIoU $\uparrow$                  & ECE $\downarrow$                  & mUnc              & mIoU $\uparrow$         & ECE $\downarrow$      & mUnc          \\ \hline
CE                      & \multicolumn{1}{c|}{RN18}                     & 69.0                  & 0.035                 & 0.088              & 71.9         & 0.025     & 0.088         \\
U-CE$_{\alpha=1}$       & \multicolumn{1}{c|}{RN18}                     & 69.5                  & 0.036                 & 0.089              & 72.6         & 0.027     & 0.088         \\
U-CE$_{\alpha=10}$      & \multicolumn{1}{c|}{RN18}                     & 71.8                  & 0.029                 & 0.085              & 73.5         & 0.018     & 0.084         \\ \hline
CE                      & \multicolumn{1}{c|}{RN101}                    & 74.6                  & 0.026                 & 0.080              & 76.3         & 0.041     & 0.076         \\
U-CE$_{\alpha=1}$       & \multicolumn{1}{c|}{RN101}                    & 74.8                  & 0.024                 & 0.079              & 77.0         & 0.041     & 0.076         \\
U-CE$_{\alpha=10}$      & \multicolumn{1}{c|}{RN101}                    & 76.6                  & 0.022                 & 0.073              & 77.7         & 0.040     & 0.073         \\
\end{tabular}
\end{adjustbox}
\end{center}
    \caption{A more detailed quantitative comparison between regular CE and U-CE on the Cityscapes dataset \cite{cordts2016CityscapesDataset} using a dropout ratio of 20\%. The provided numbers represent the mIoU $\uparrow$, ECE $\downarrow$, and mUnc.}
    \label{tab:city_detailed}
\end{table}

\begin{figure*}[ht!]
    \centering
    \begin{subfigure}{0.025\textwidth}
        \rotatebox[origin=c]{90}{CE}\quad
    \end{subfigure}\hfill
    \begin{subfigure}{0.19\textwidth}
        \includegraphics[width=\textwidth,valign=c]{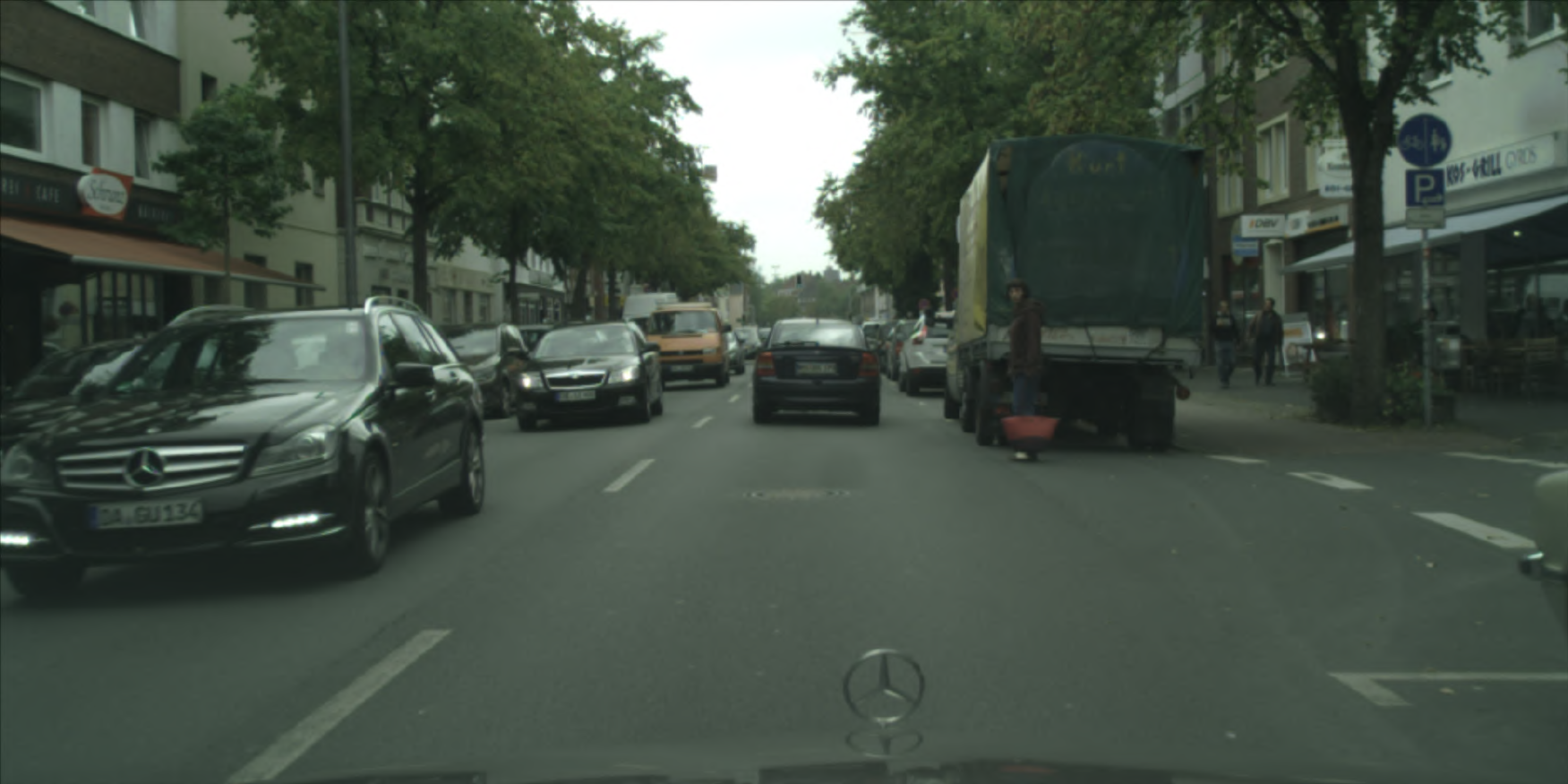}
    \end{subfigure}\hfill
    \begin{subfigure}{0.19\textwidth}
        \includegraphics[width=\textwidth,valign=c]{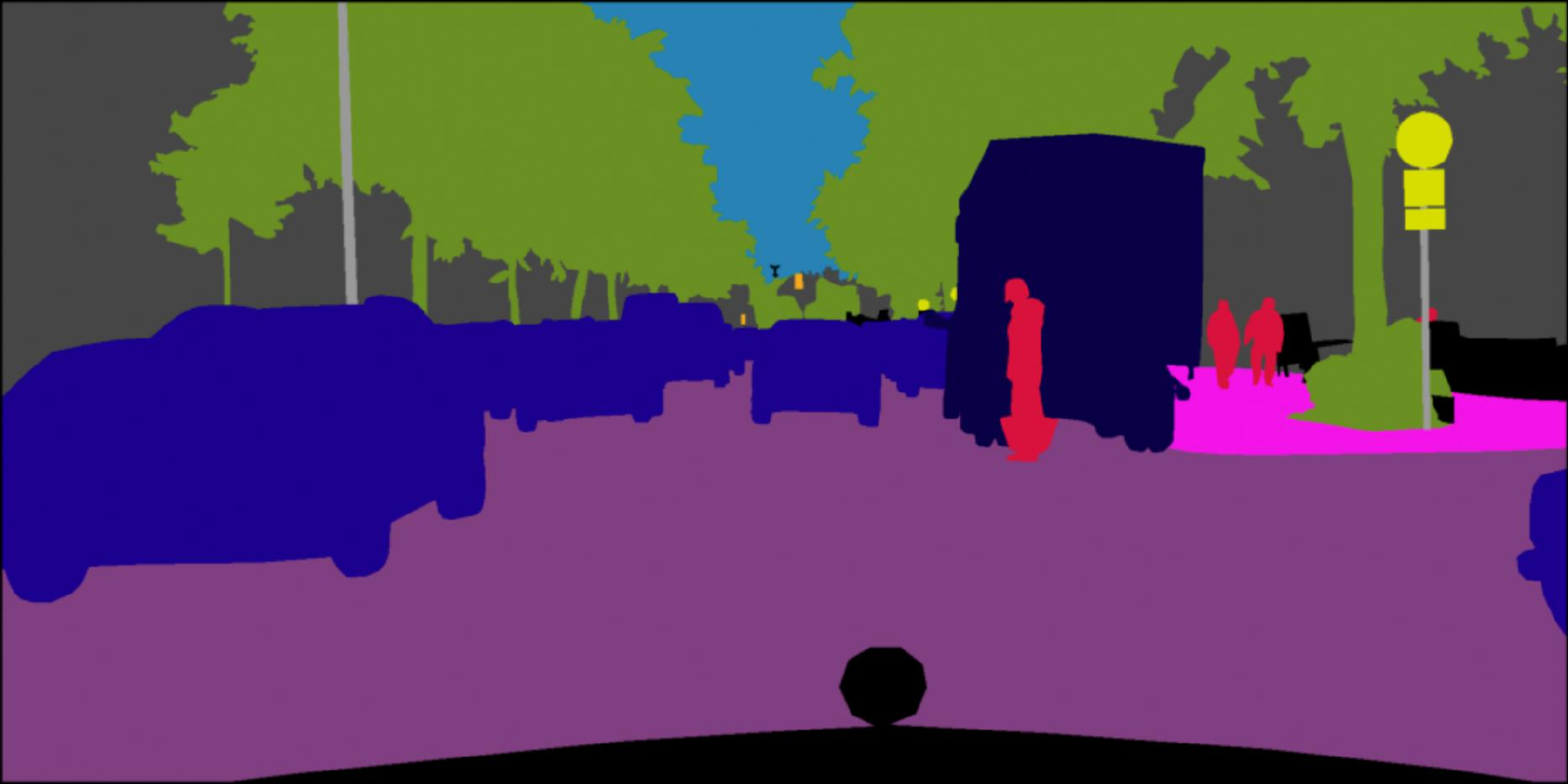}
    \end{subfigure}\hfill
    \begin{subfigure}{0.19\textwidth}
        \includegraphics[width=\textwidth,valign=c]{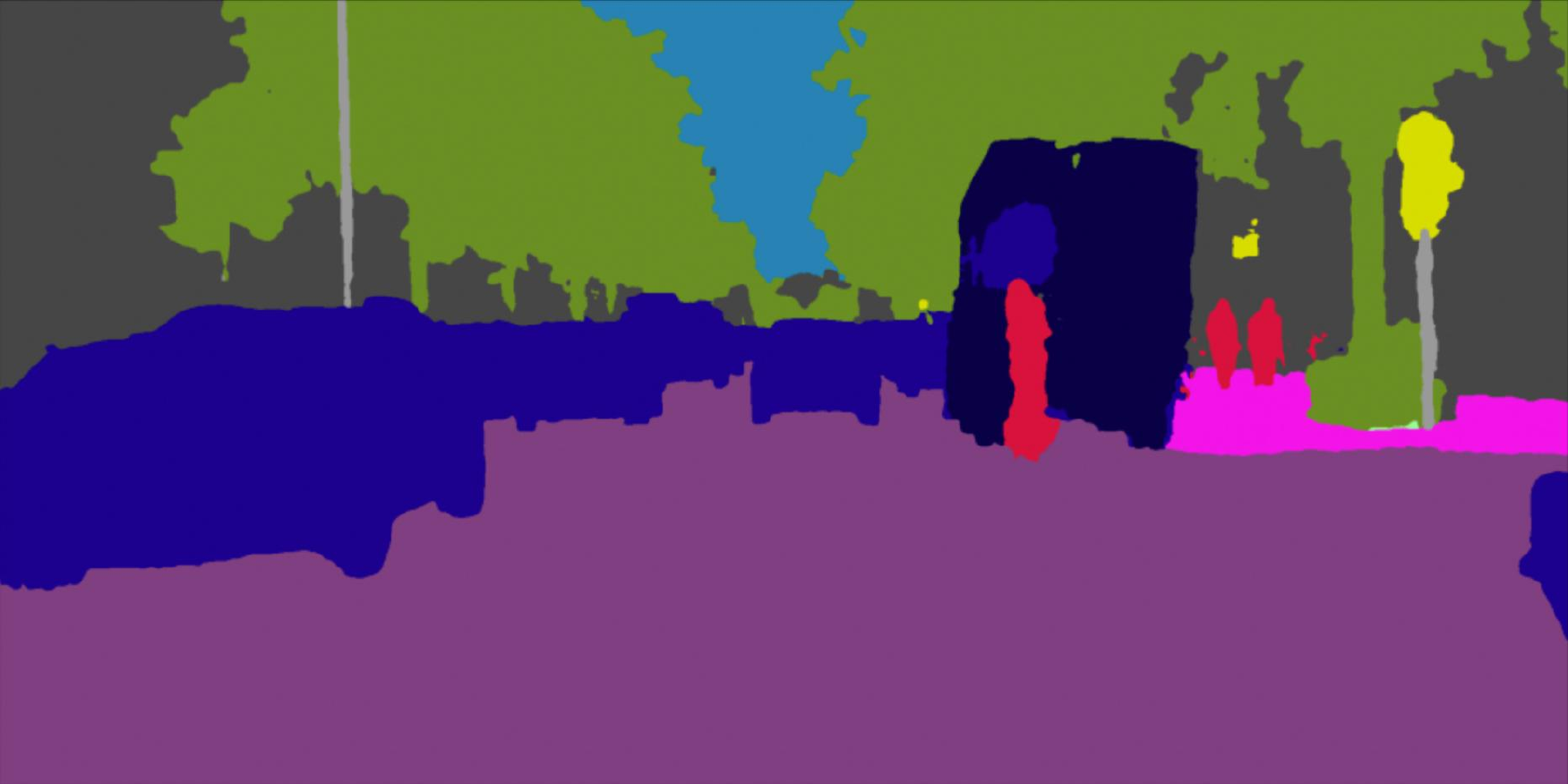}
    \end{subfigure}\hfill
    \begin{subfigure}{0.19\textwidth}
        \includegraphics[width=\textwidth,valign=c]{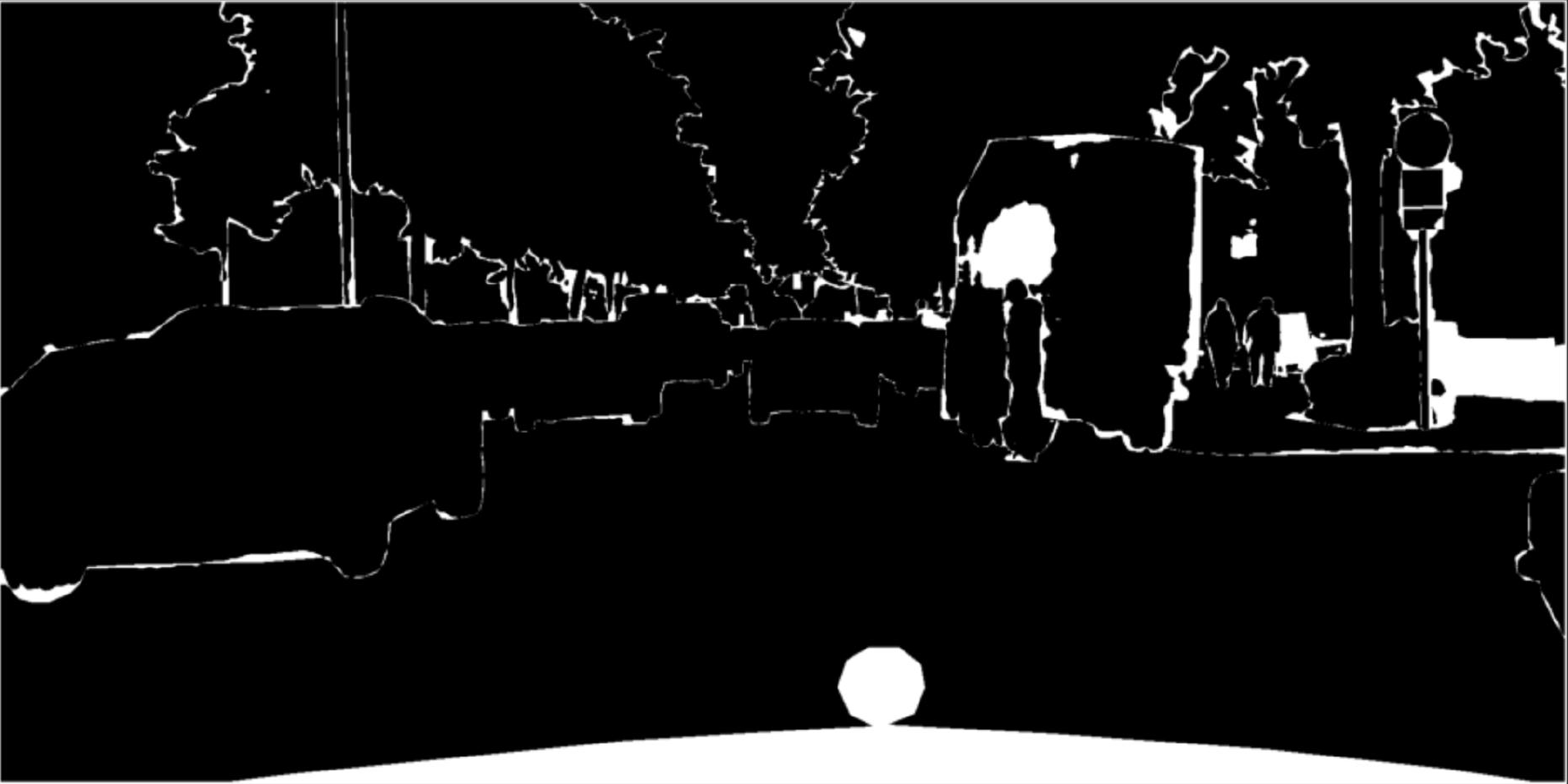}
    \end{subfigure}\hfill
    \begin{subfigure}{0.19\textwidth}
        \includegraphics[width=\textwidth,valign=c]{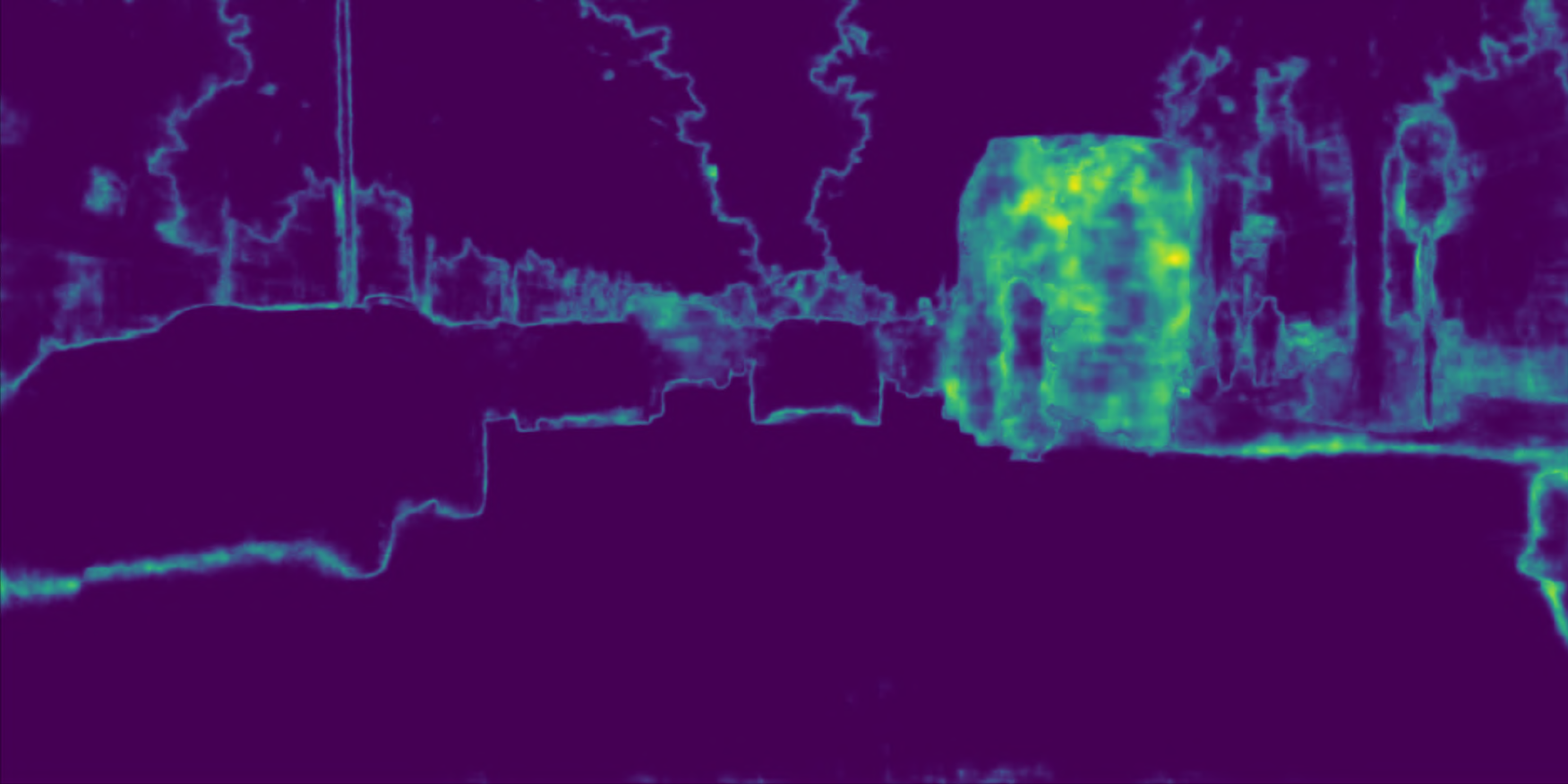}
    \end{subfigure}

    \begin{subfigure}{0.025\textwidth}
        \rotatebox[origin=c]{90}{U-CE$_{\alpha=1}$}\quad
    \end{subfigure}\hfill
    \begin{subfigure}{0.19\textwidth}
        \includegraphics[width=\textwidth,valign=c]{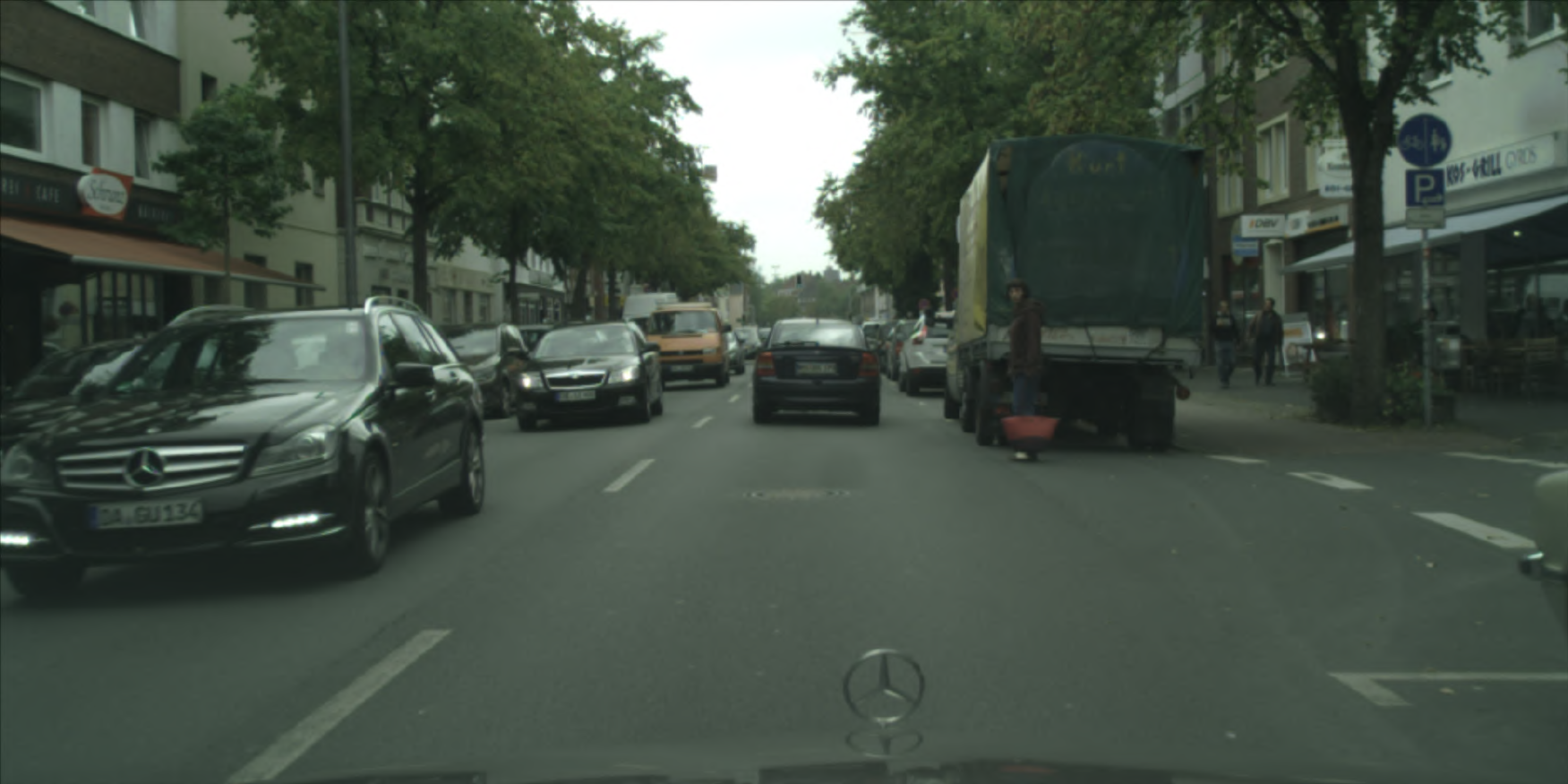}
    \end{subfigure}\hfill
    \begin{subfigure}{0.19\textwidth}
        \includegraphics[width=\textwidth,valign=c]{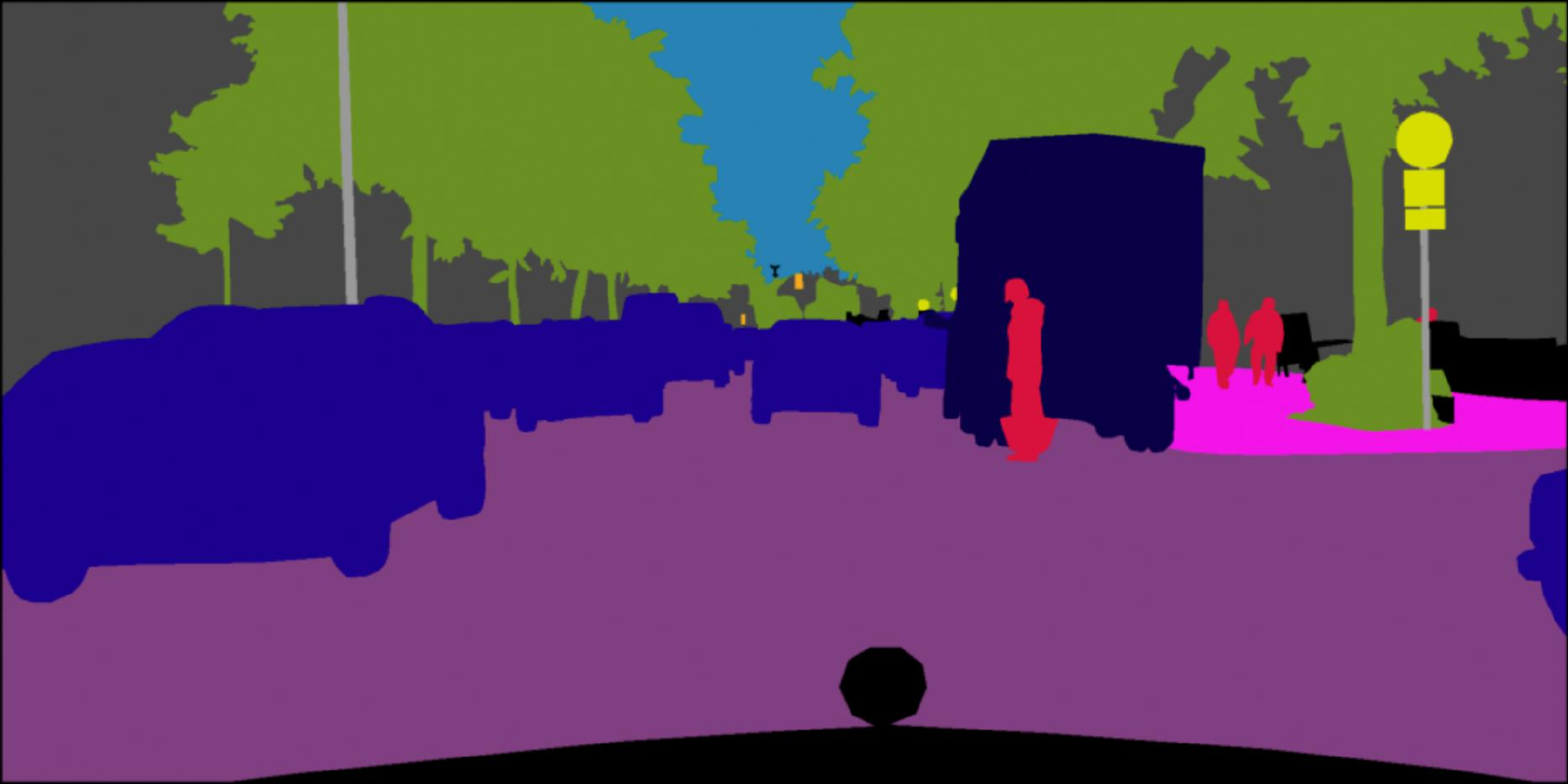}
    \end{subfigure}\hfill
    \begin{subfigure}{0.19\textwidth}
        \includegraphics[width=\textwidth,valign=c]{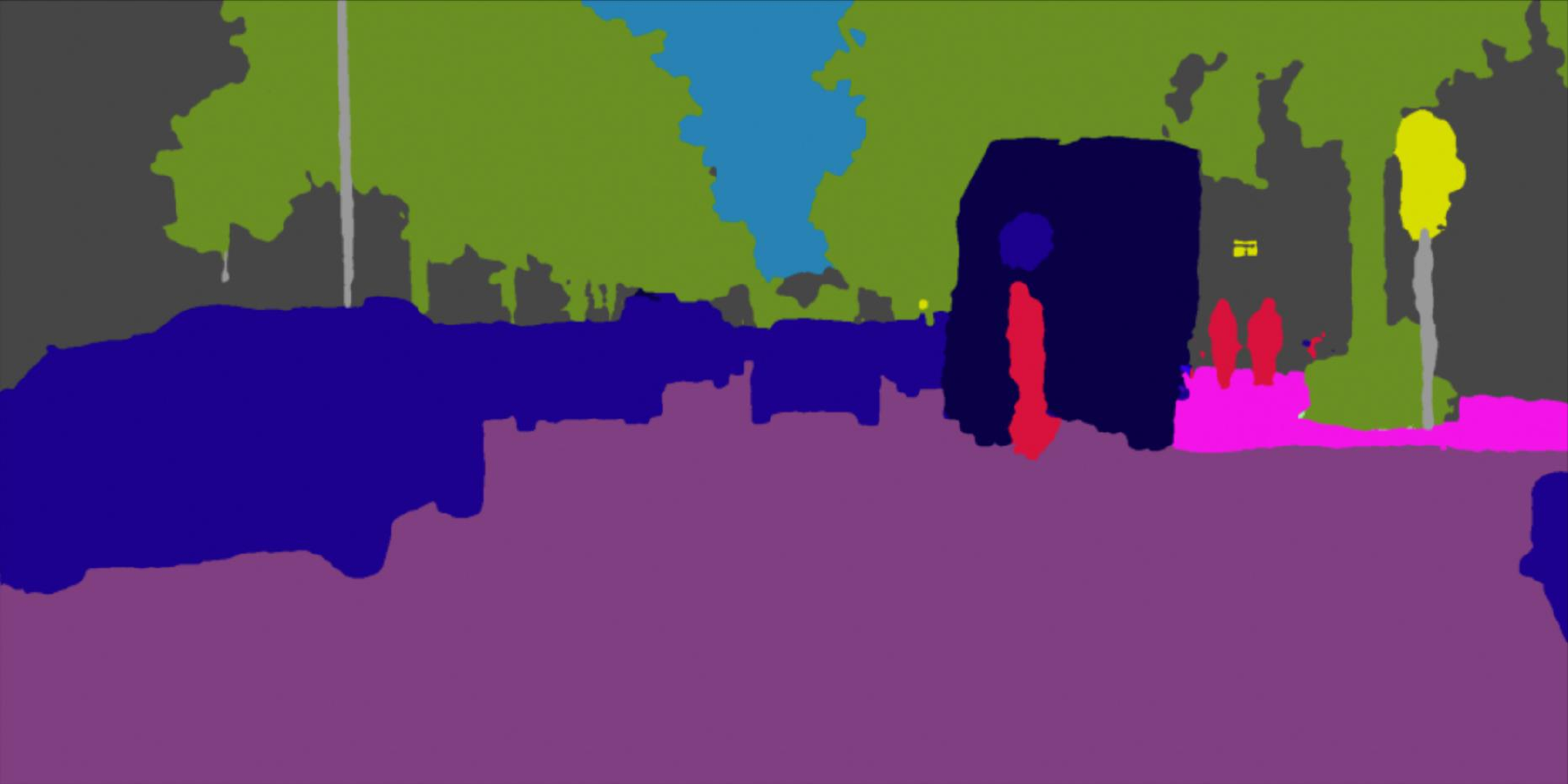}
    \end{subfigure}\hfill
    \begin{subfigure}{0.19\textwidth}
        \includegraphics[width=\textwidth,valign=c]{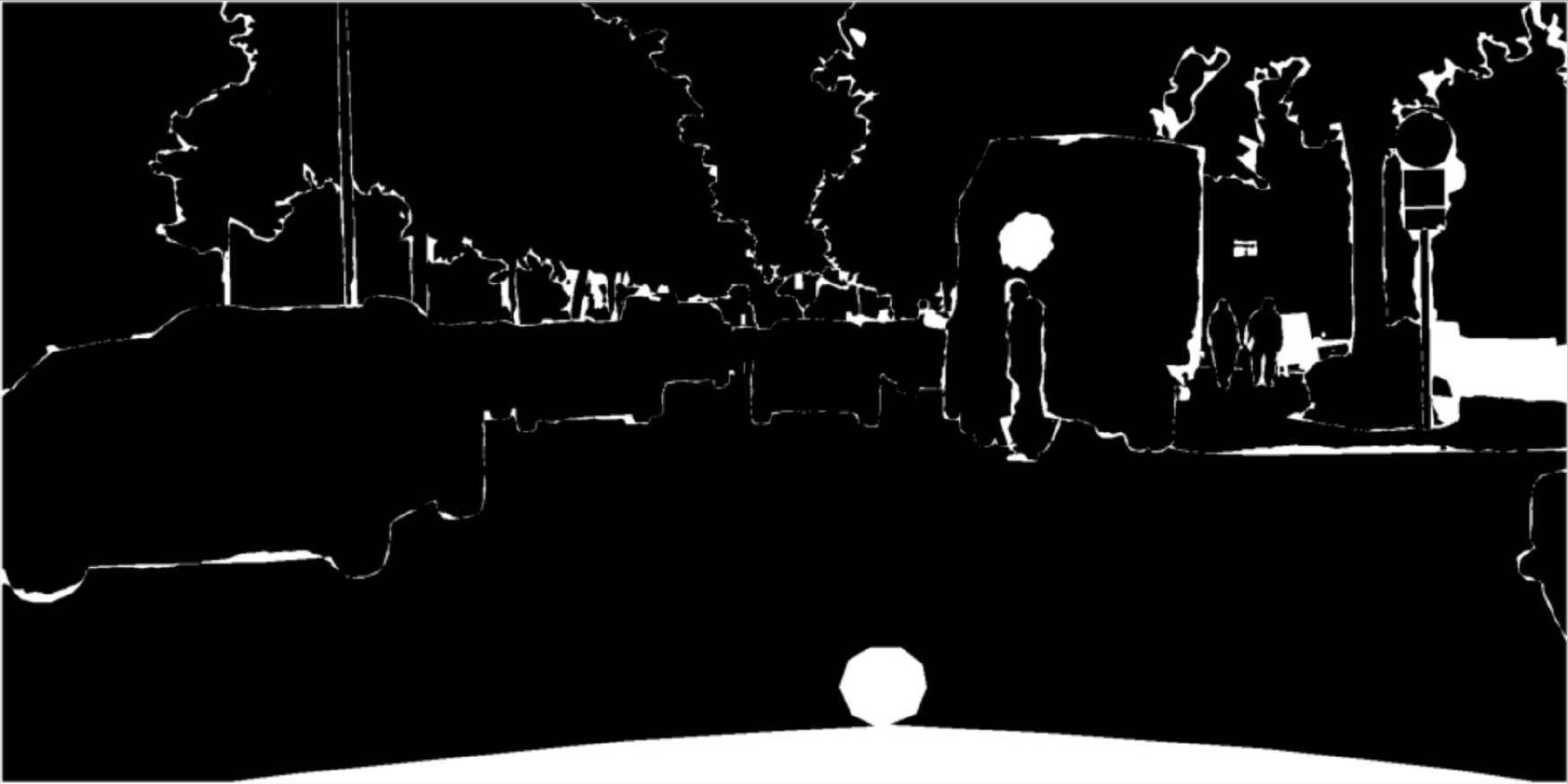}
    \end{subfigure}\hfill
    \begin{subfigure}{0.19\textwidth}
        \includegraphics[width=\textwidth,valign=c]{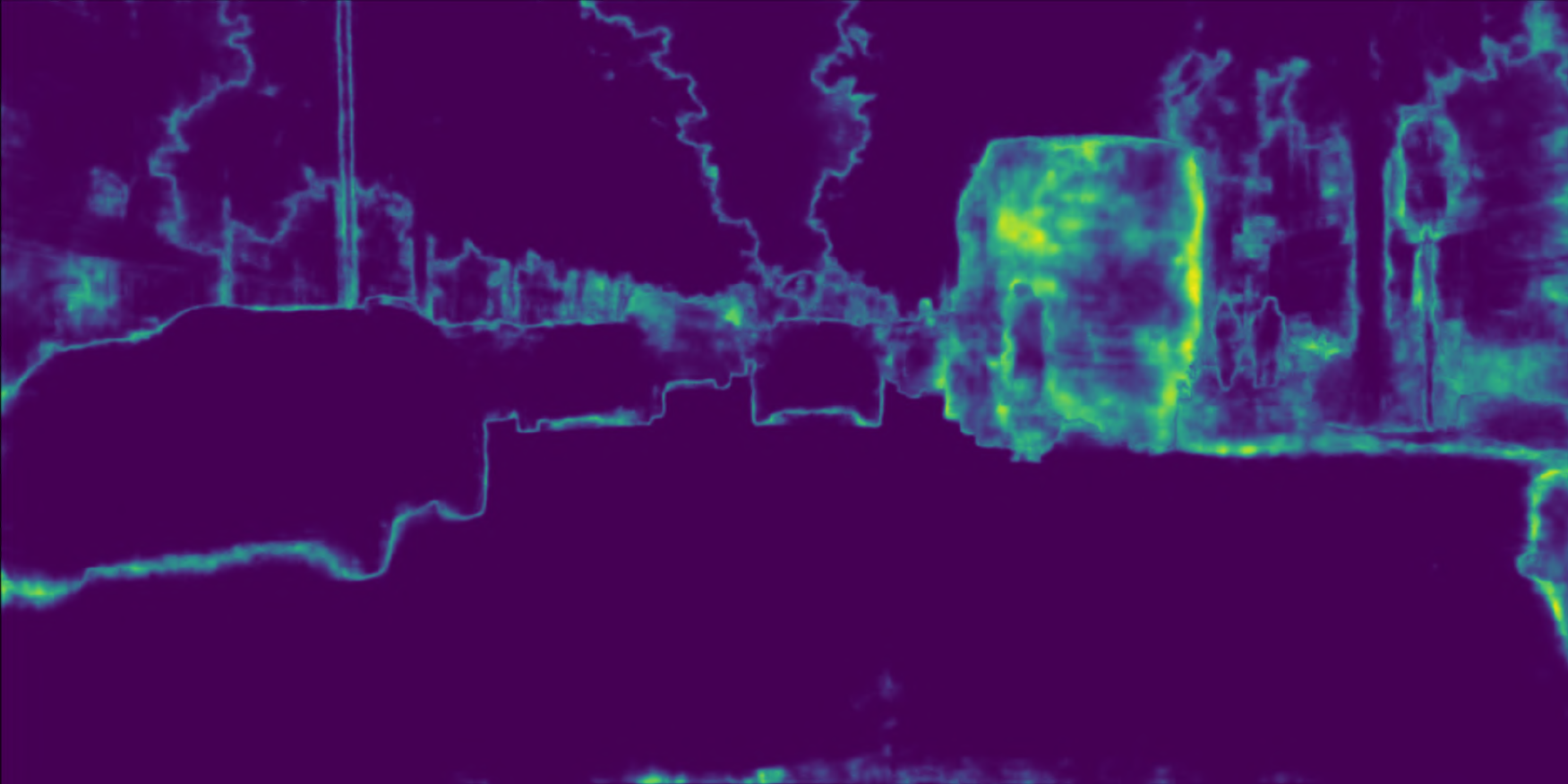}
    \end{subfigure}

    \begin{subfigure}{0.025\textwidth}
        \rotatebox[origin=c]{90}{U-CE$_{\alpha=10}$}\quad
    \end{subfigure}\hfill
    \begin{subfigure}{0.19\textwidth}
        \includegraphics[width=\textwidth,valign=c]{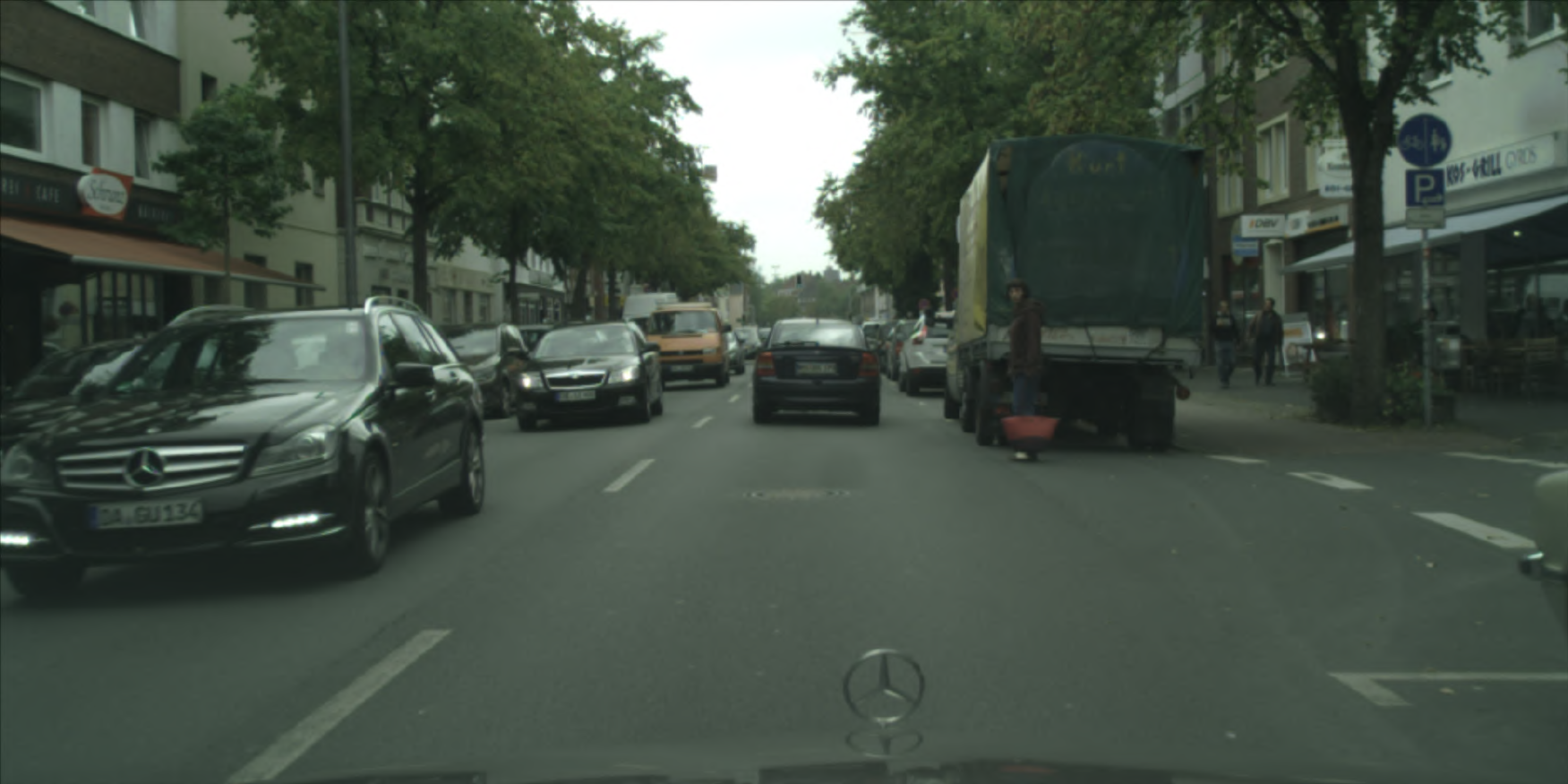}
    \end{subfigure}\hfill
    \begin{subfigure}{0.19\textwidth}
        \includegraphics[width=\textwidth,valign=c]{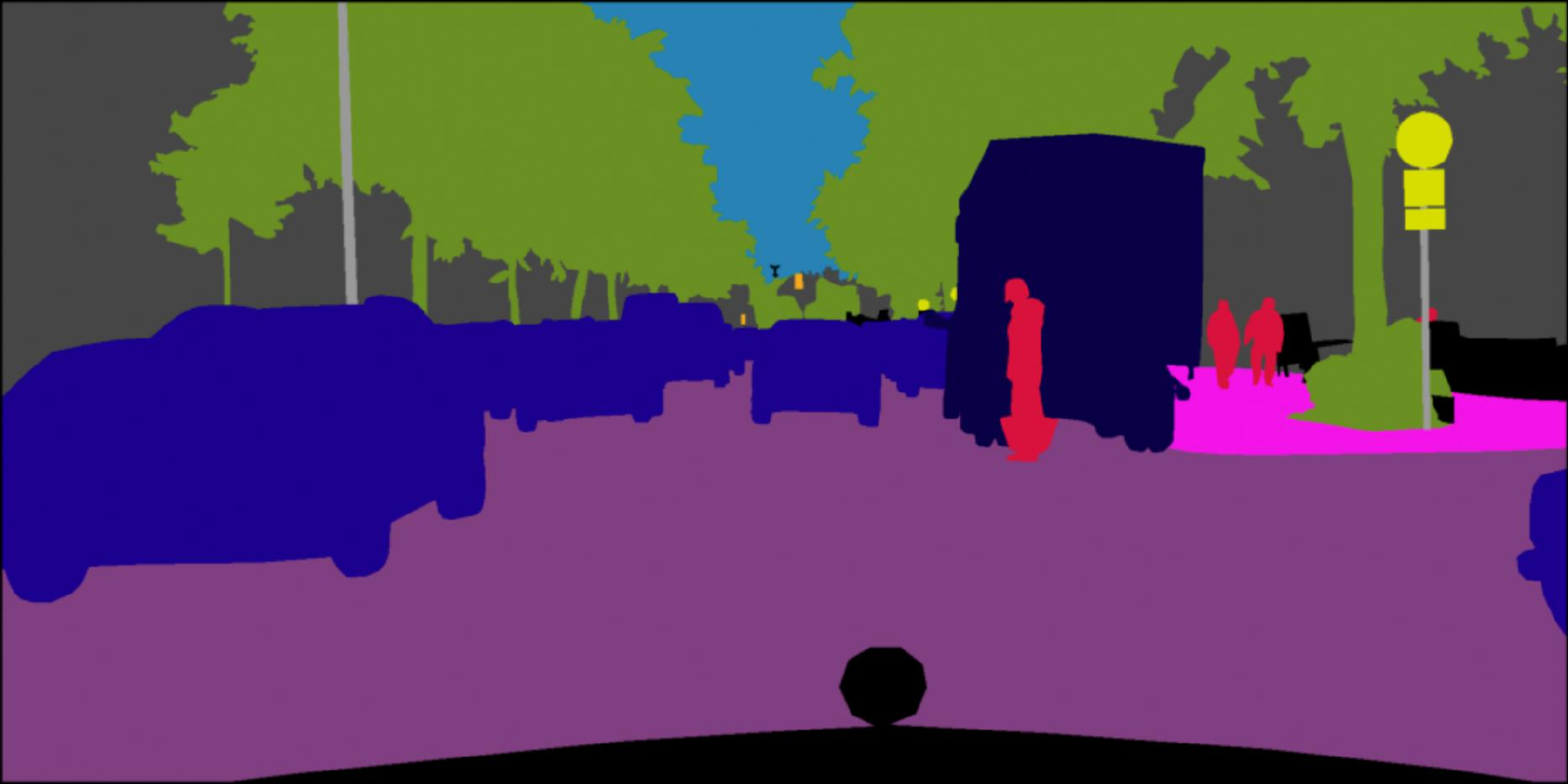}
    \end{subfigure}\hfill
    \begin{subfigure}{0.19\textwidth}
        \includegraphics[width=\textwidth,valign=c]{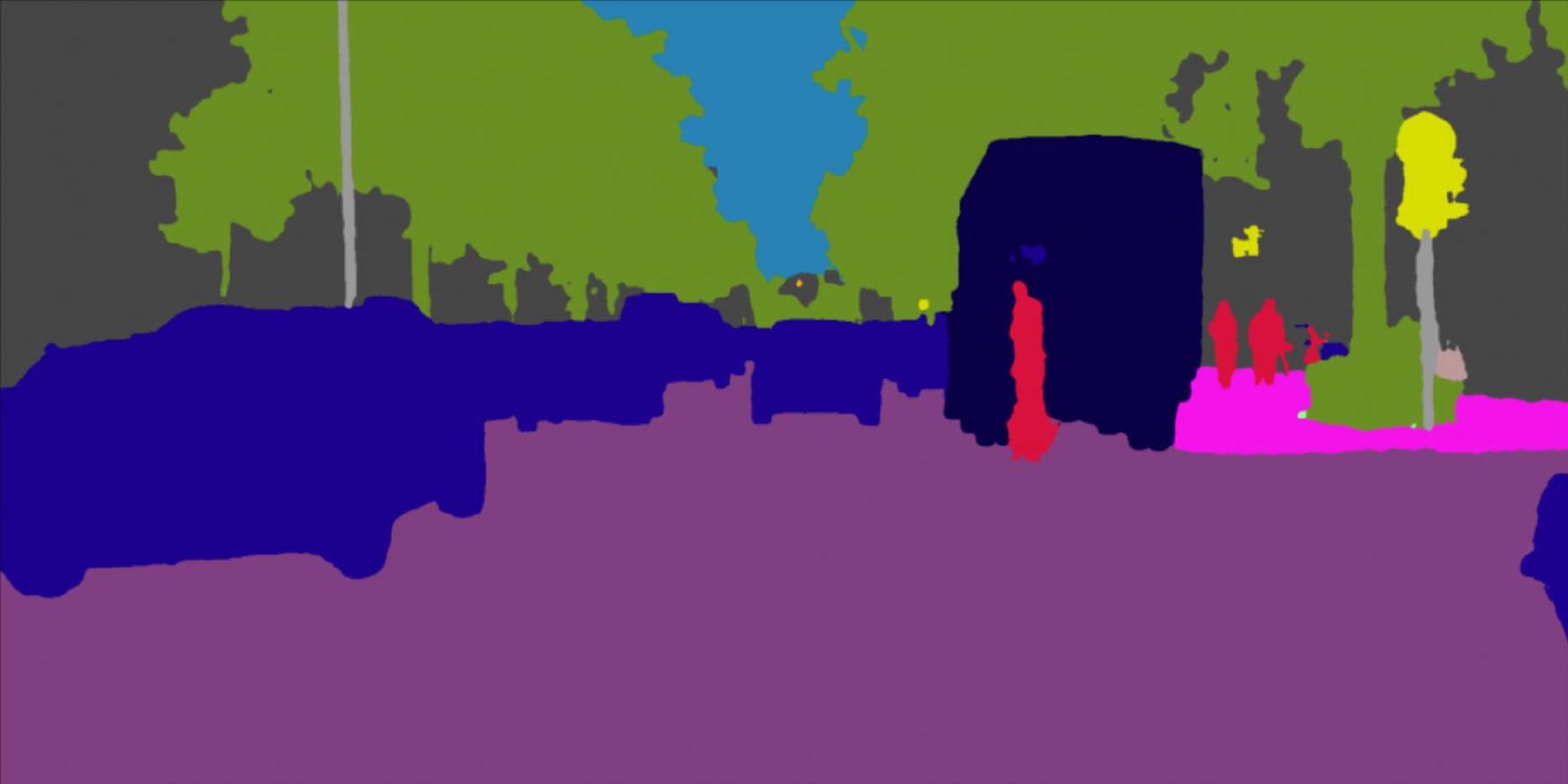}
    \end{subfigure}\hfill
    \begin{subfigure}{0.19\textwidth}
        \includegraphics[width=\textwidth,valign=c]{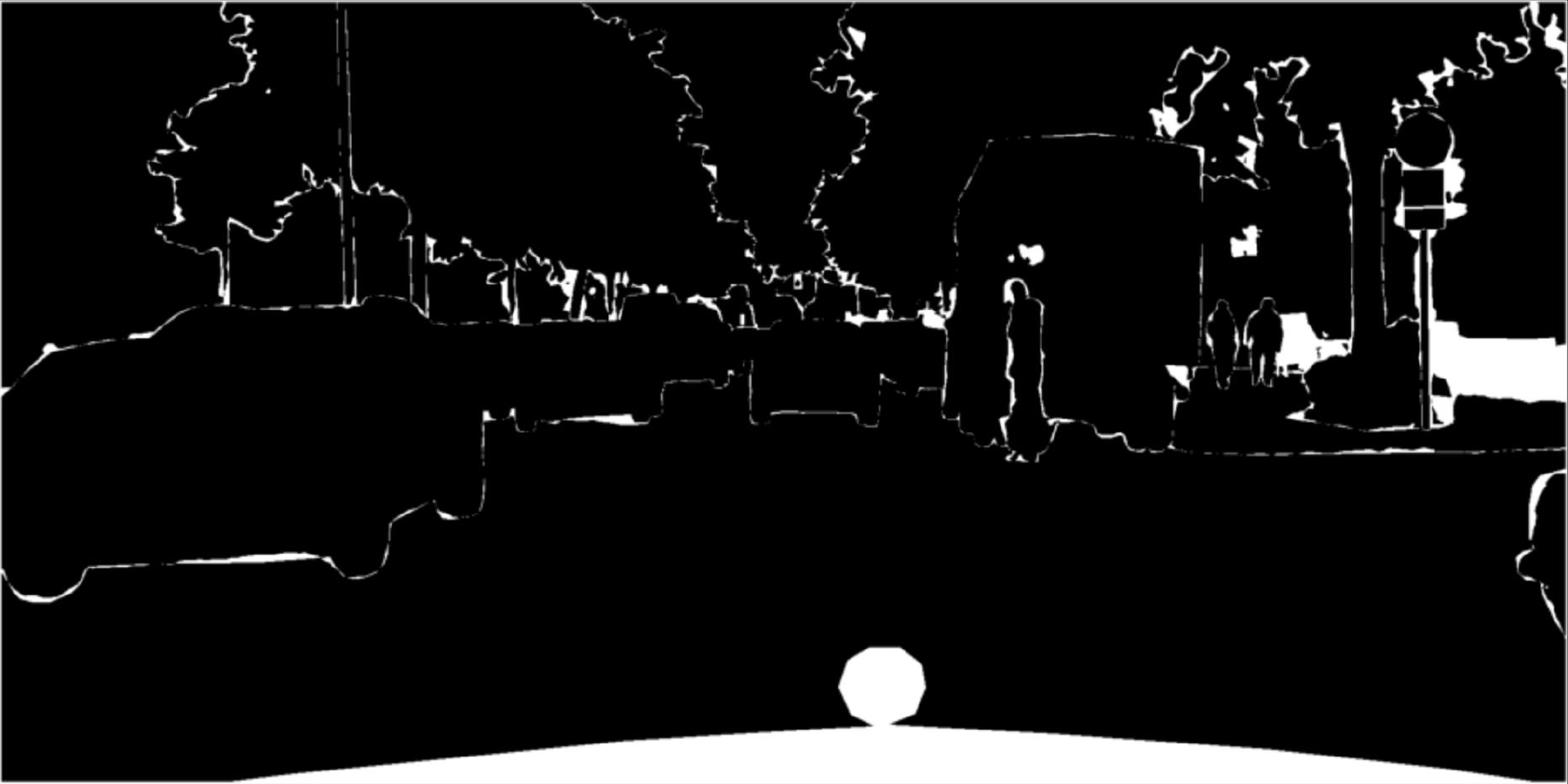}
    \end{subfigure}\hfill
    \begin{subfigure}{0.19\textwidth}
        \includegraphics[width=\textwidth,valign=c]{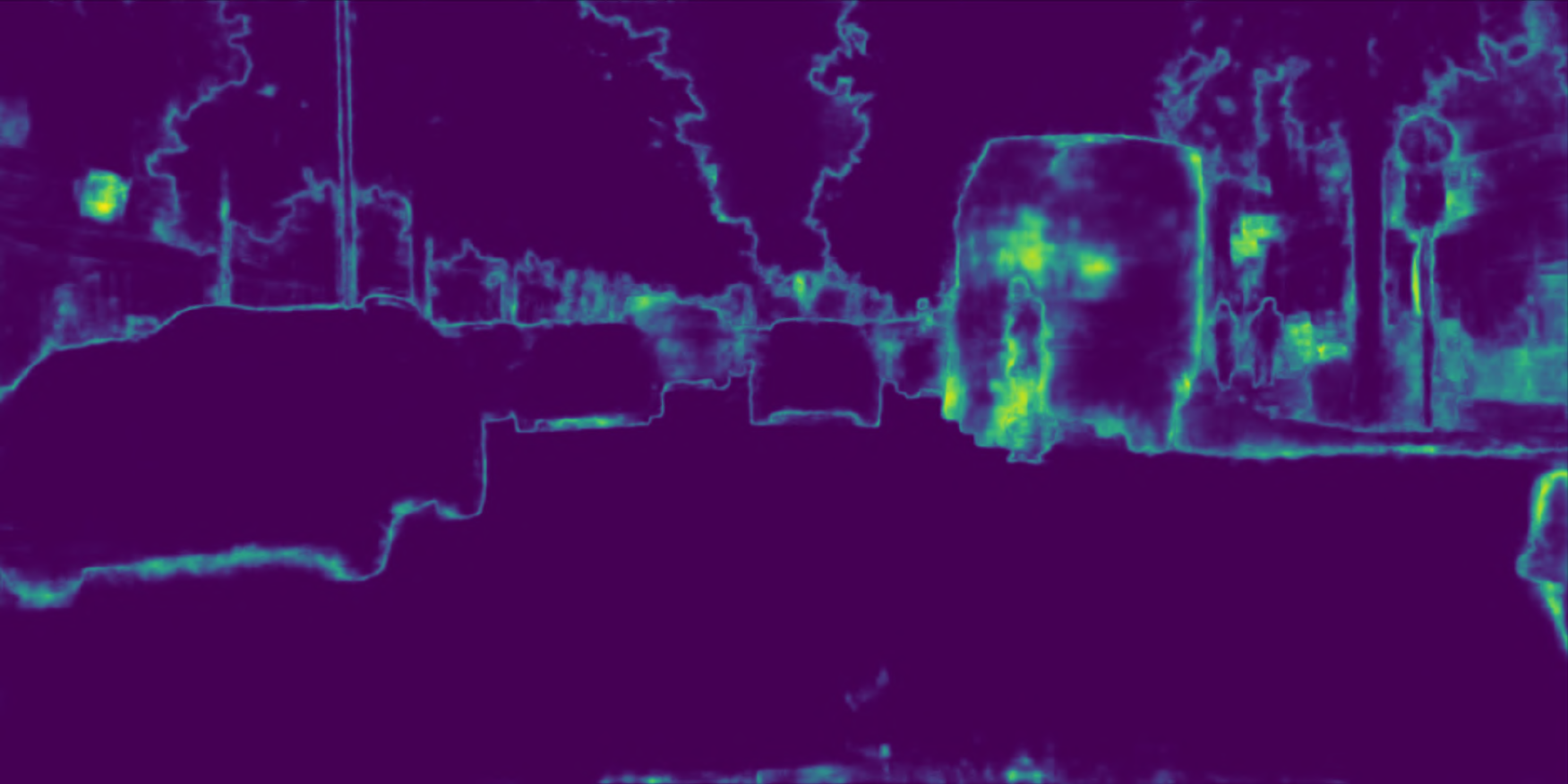}
    \end{subfigure}

    \begin{subfigure}{0.025\textwidth}
        \rotatebox[origin=c]{90}{CE}\quad
    \end{subfigure}\hfill
    \begin{subfigure}{0.19\textwidth}
        \includegraphics[width=\textwidth,valign=c]{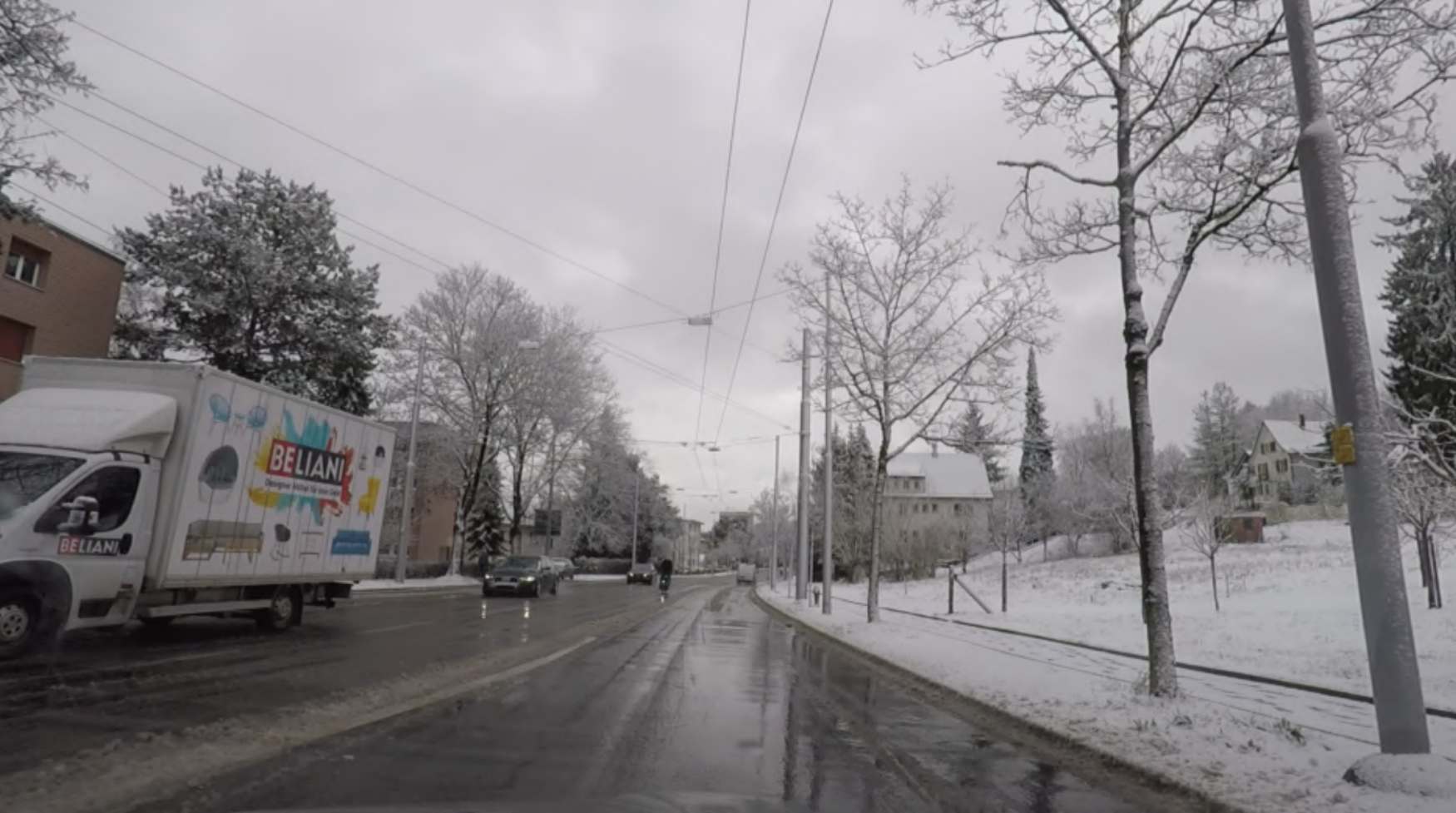}
    \end{subfigure}\hfill
    \begin{subfigure}{0.19\textwidth}
        \includegraphics[width=\textwidth,valign=c]{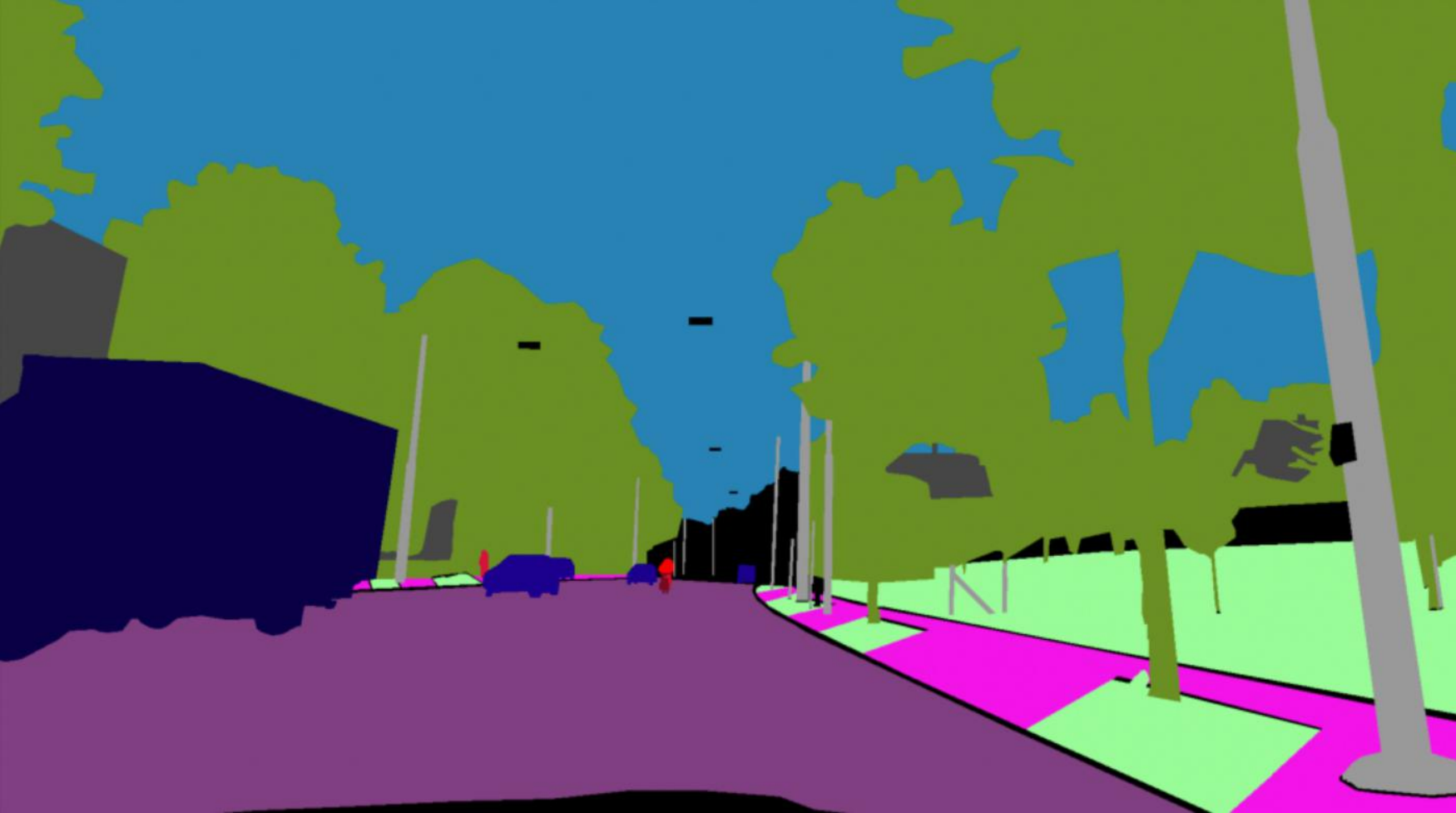}
    \end{subfigure}\hfill
    \begin{subfigure}{0.19\textwidth}
        \includegraphics[width=\textwidth,valign=c]{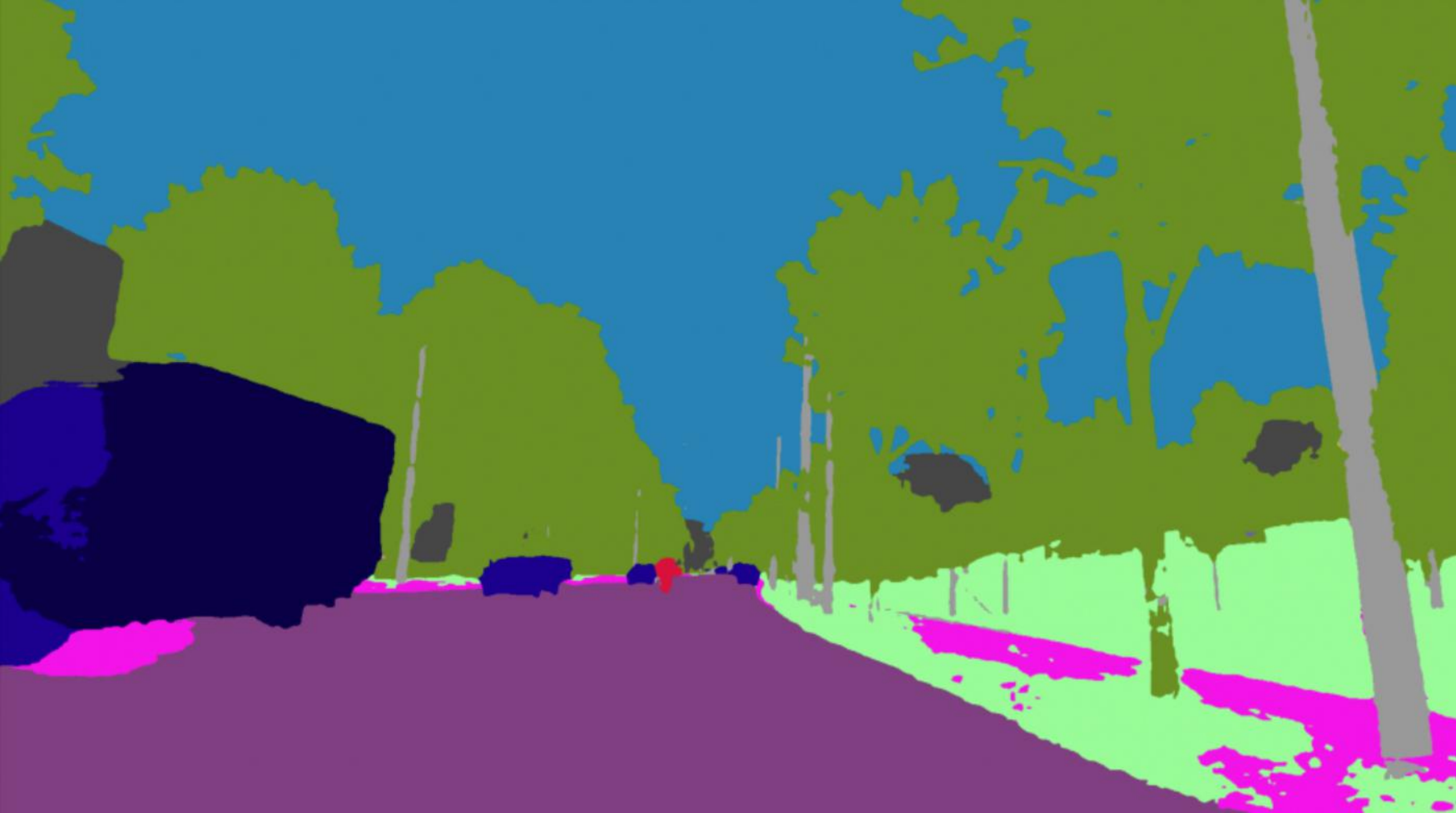}
    \end{subfigure}\hfill
    \begin{subfigure}{0.19\textwidth}
        \includegraphics[width=\textwidth,valign=c]{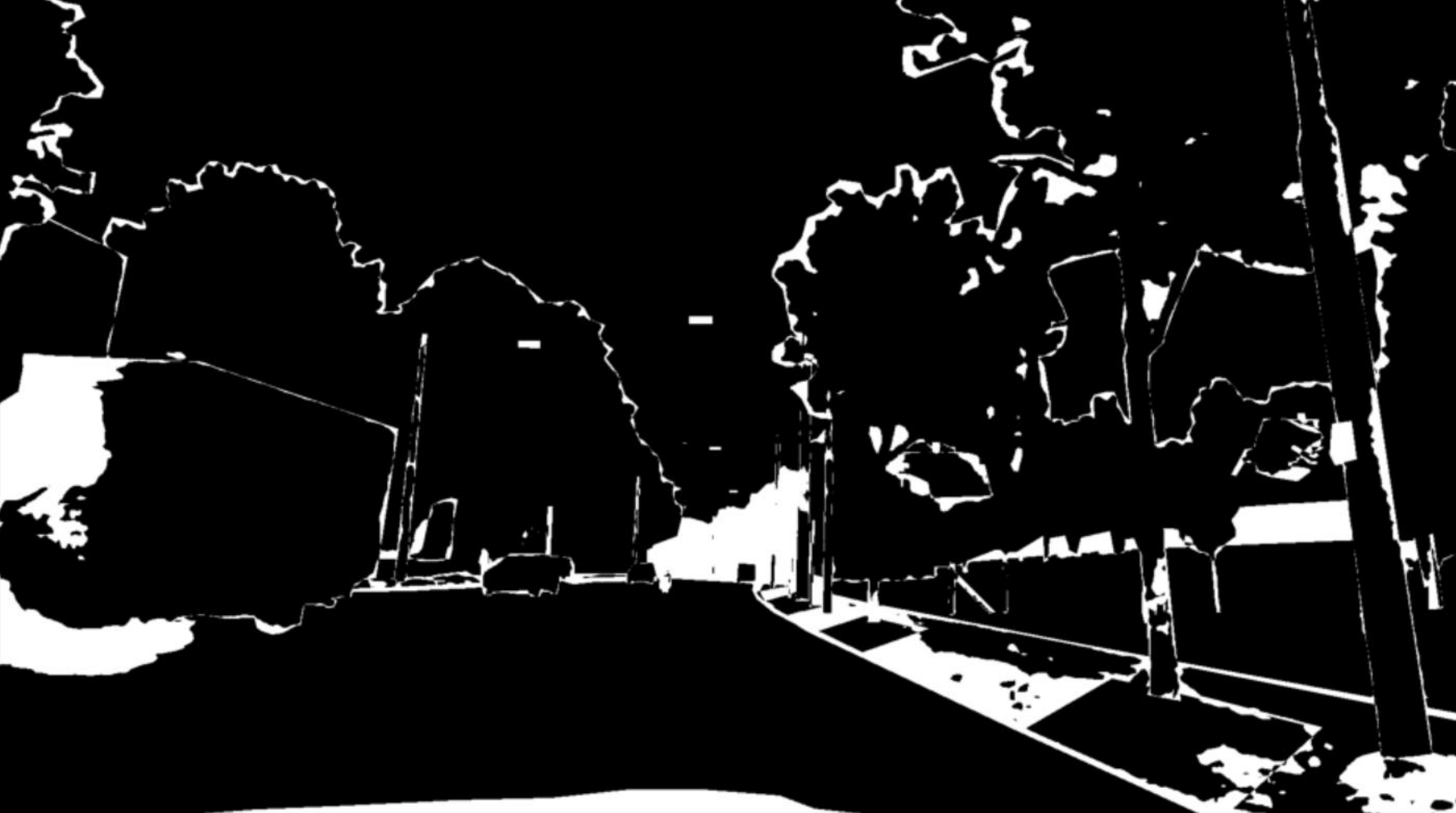}
    \end{subfigure}\hfill
    \begin{subfigure}{0.19\textwidth}
        \includegraphics[width=\textwidth,valign=c]{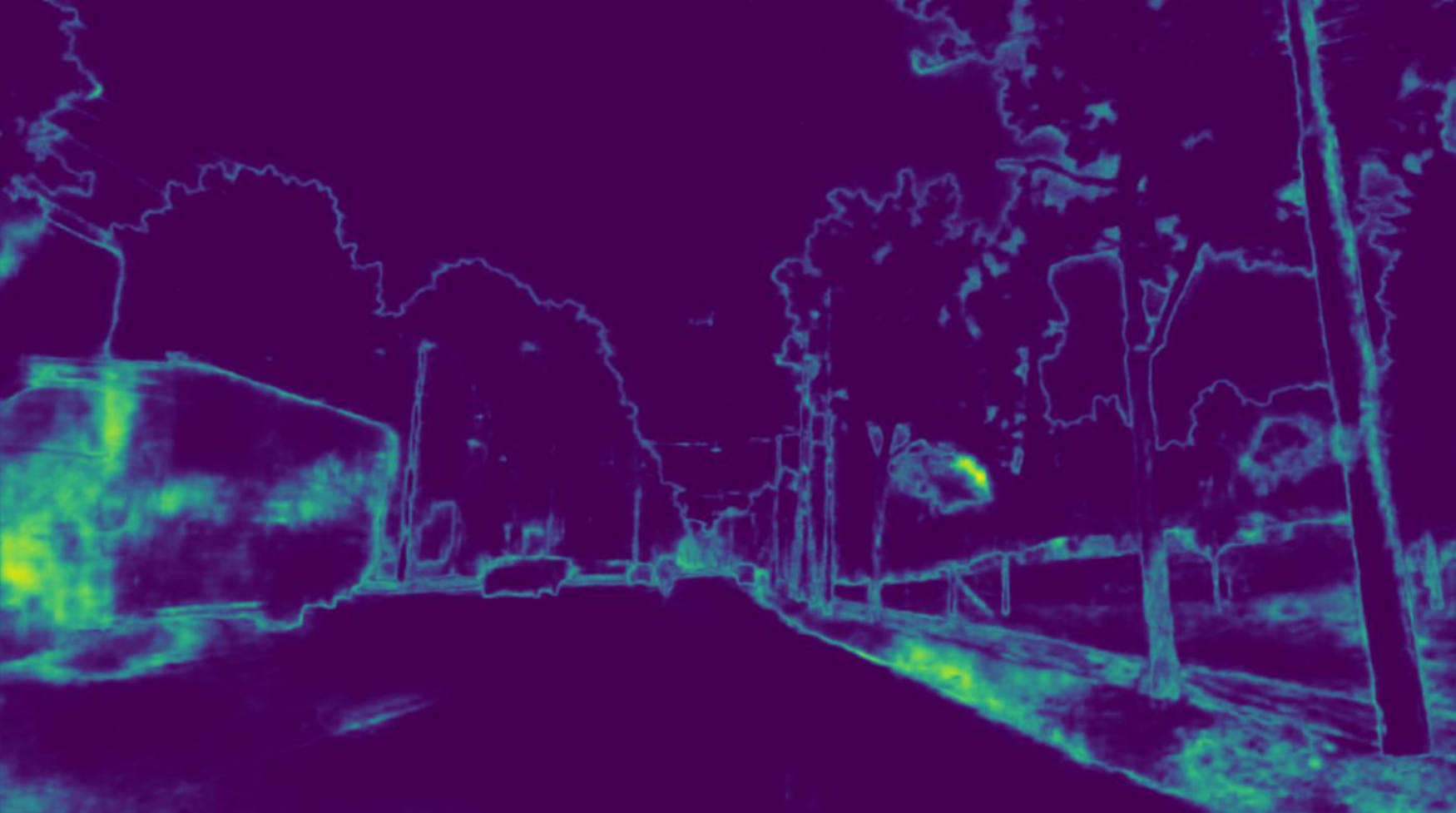}
    \end{subfigure}

    \begin{subfigure}{0.025\textwidth}
        \rotatebox[origin=c]{90}{U-CE$_{\alpha=1}$}\quad
    \end{subfigure}\hfill
    \begin{subfigure}{0.19\textwidth}
        \includegraphics[width=\textwidth,valign=c]{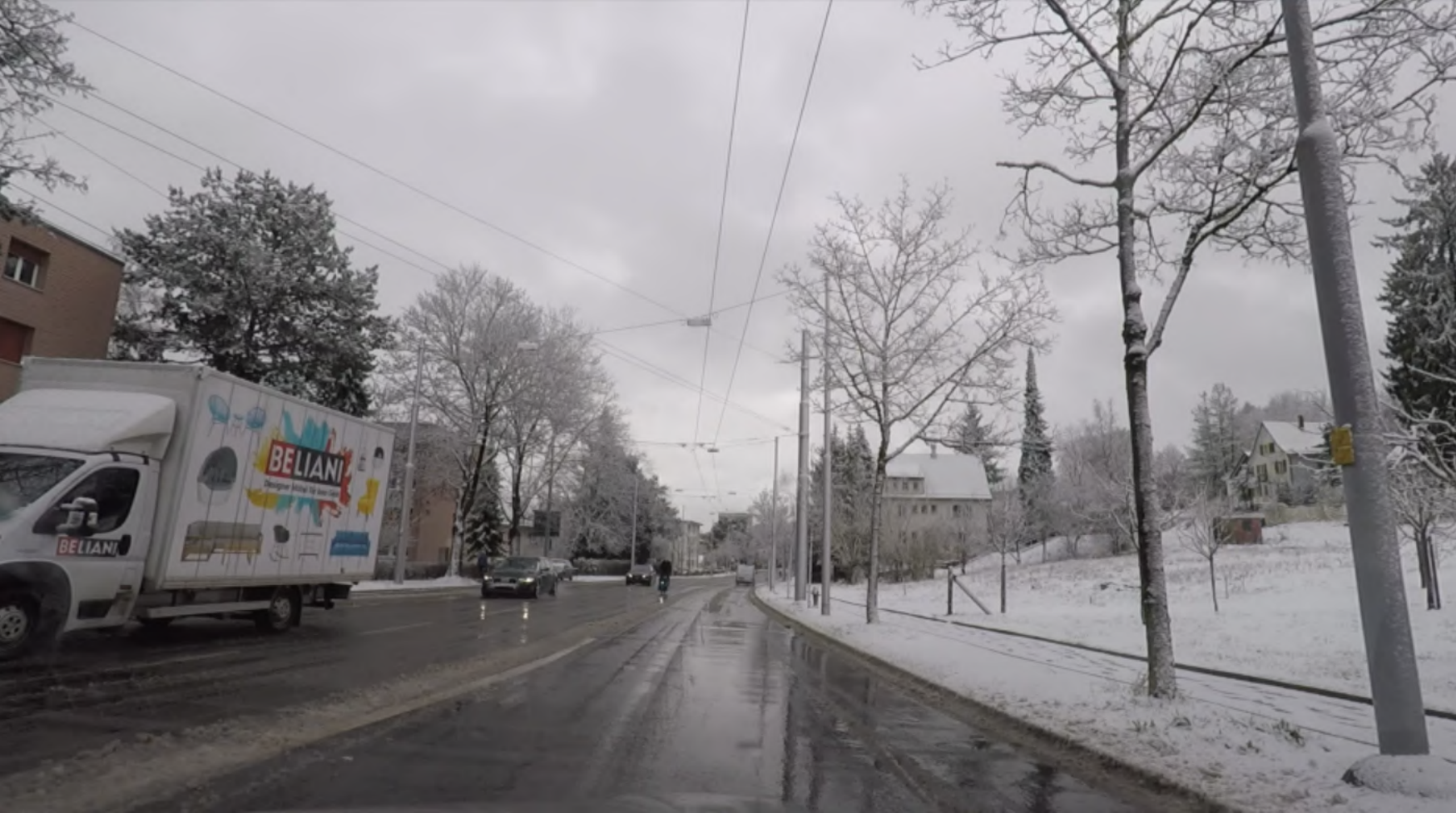}
    \end{subfigure}\hfill
    \begin{subfigure}{0.19\textwidth}
        \includegraphics[width=\textwidth,valign=c]{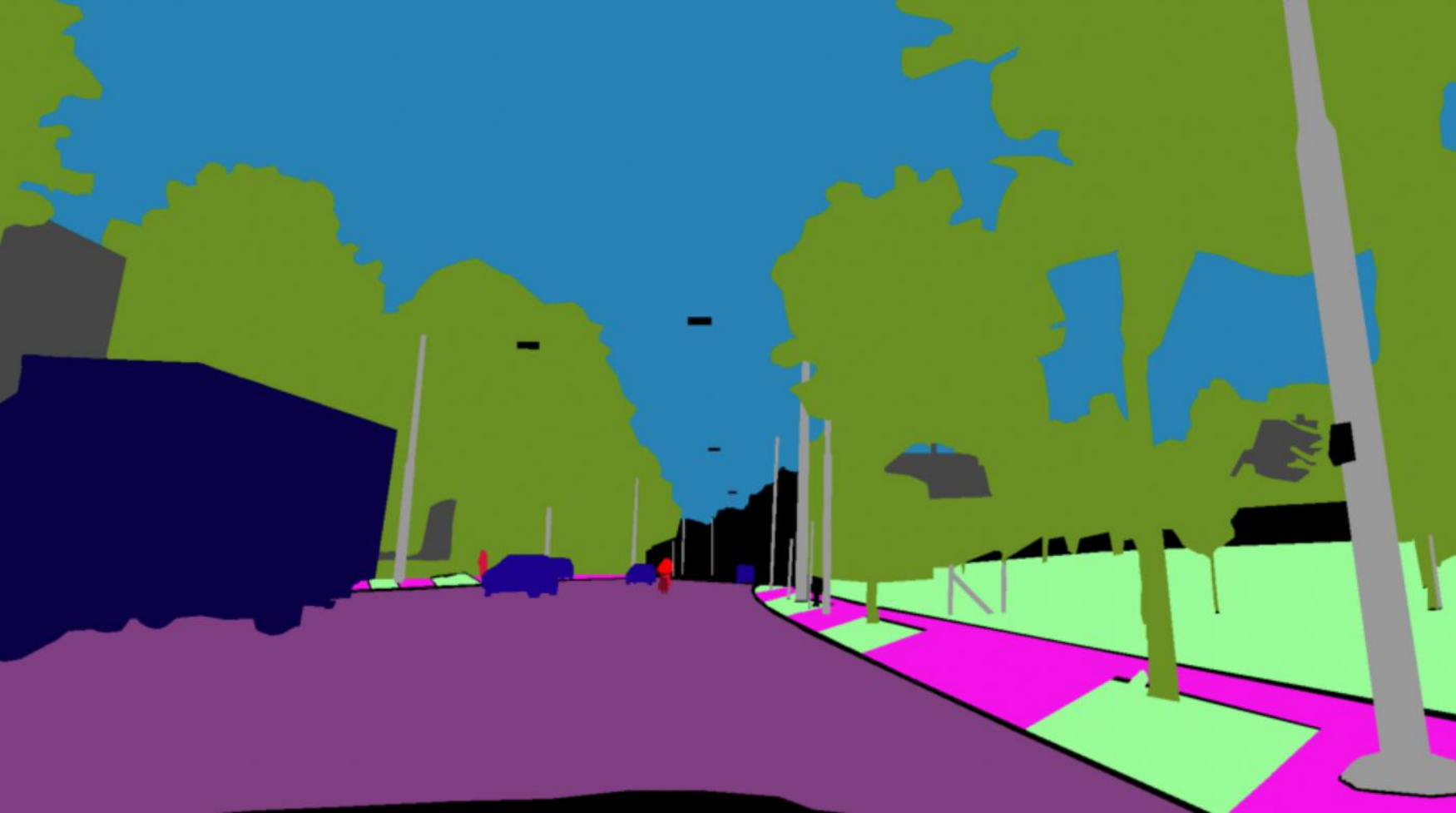}
    \end{subfigure}\hfill
    \begin{subfigure}{0.19\textwidth}
        \includegraphics[width=\textwidth,valign=c]{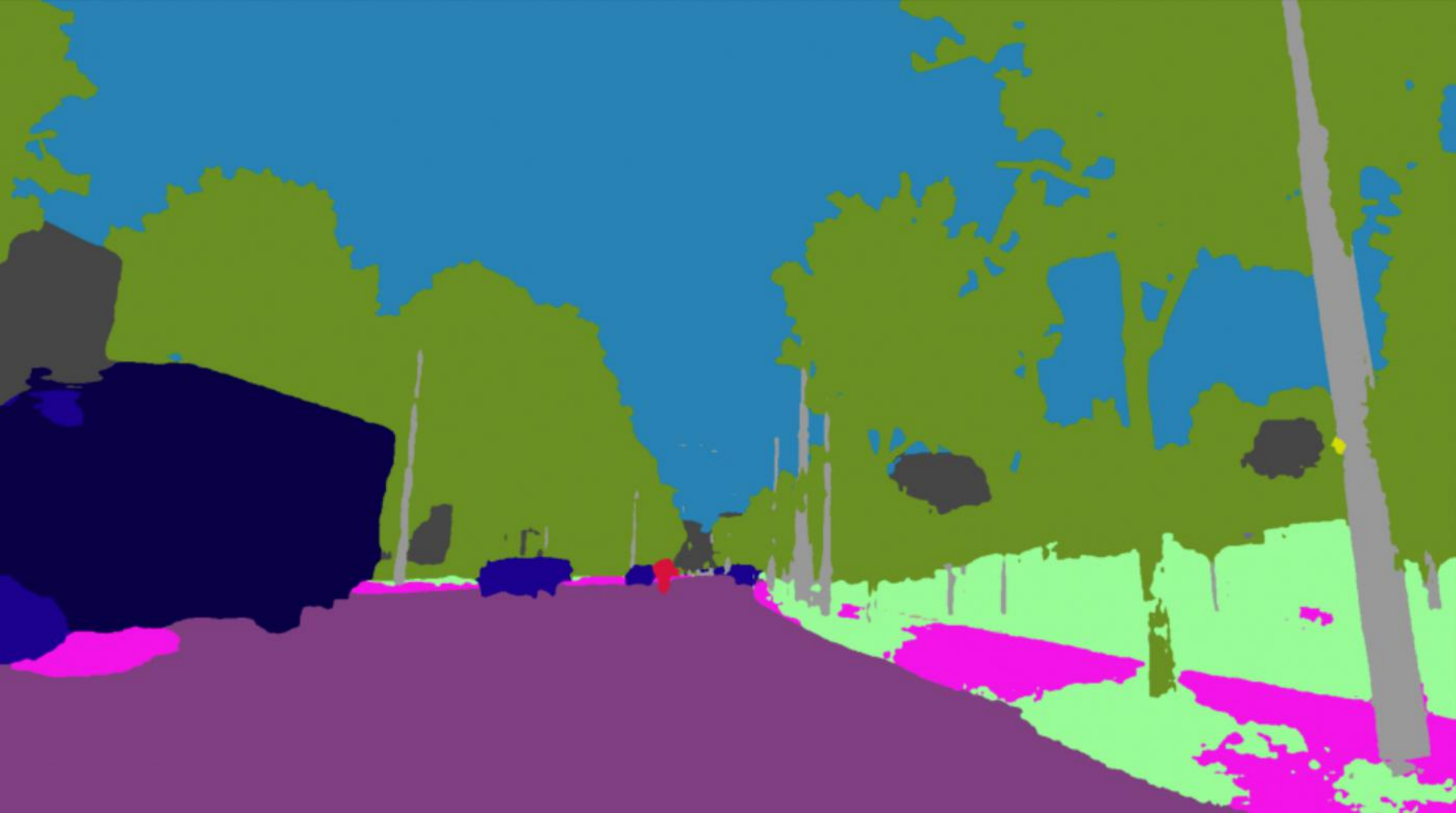}
    \end{subfigure}\hfill
    \begin{subfigure}{0.19\textwidth}
        \includegraphics[width=\textwidth,valign=c]{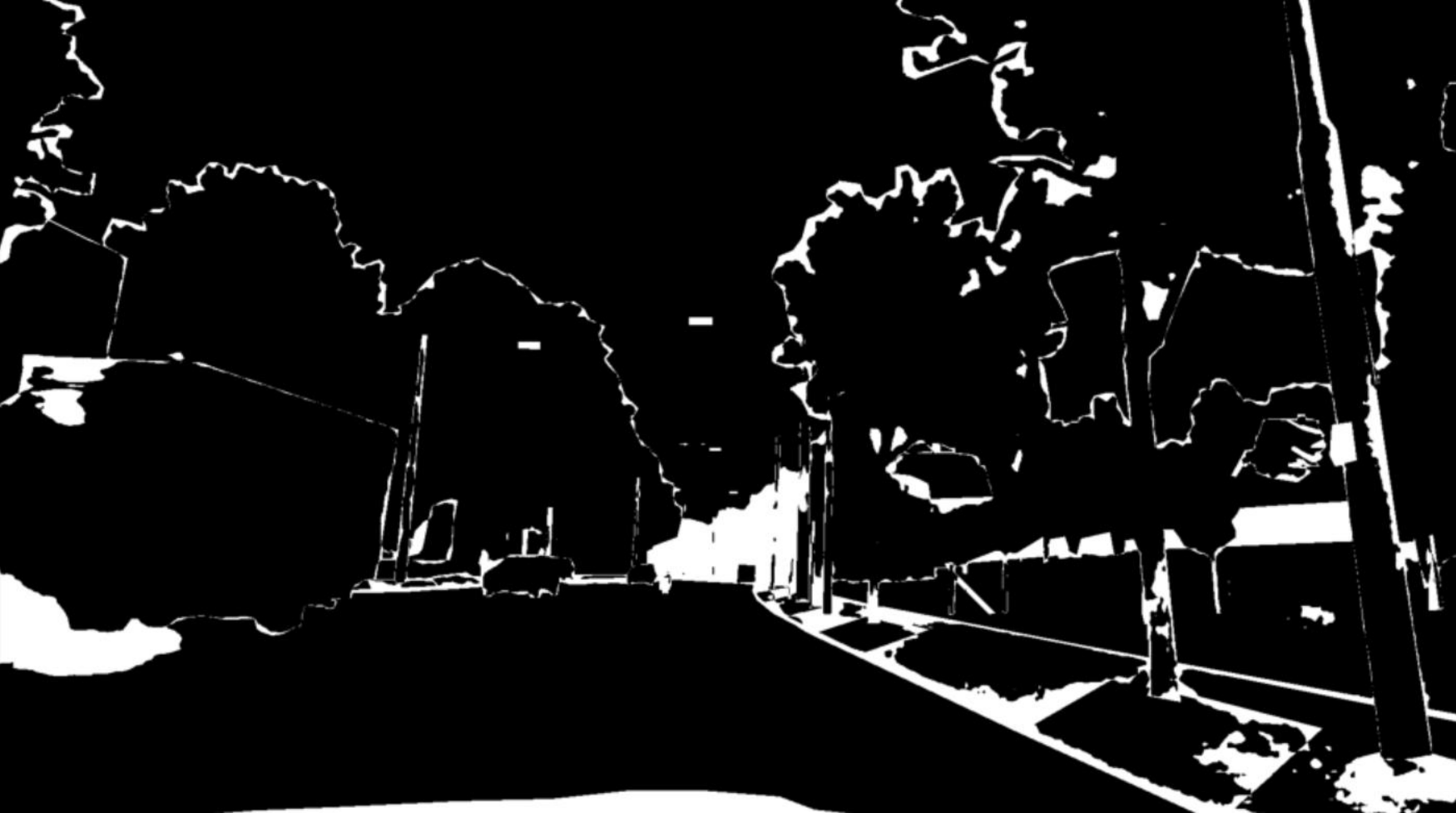}
    \end{subfigure}\hfill
    \begin{subfigure}{0.19\textwidth}
        \includegraphics[width=\textwidth,valign=c]{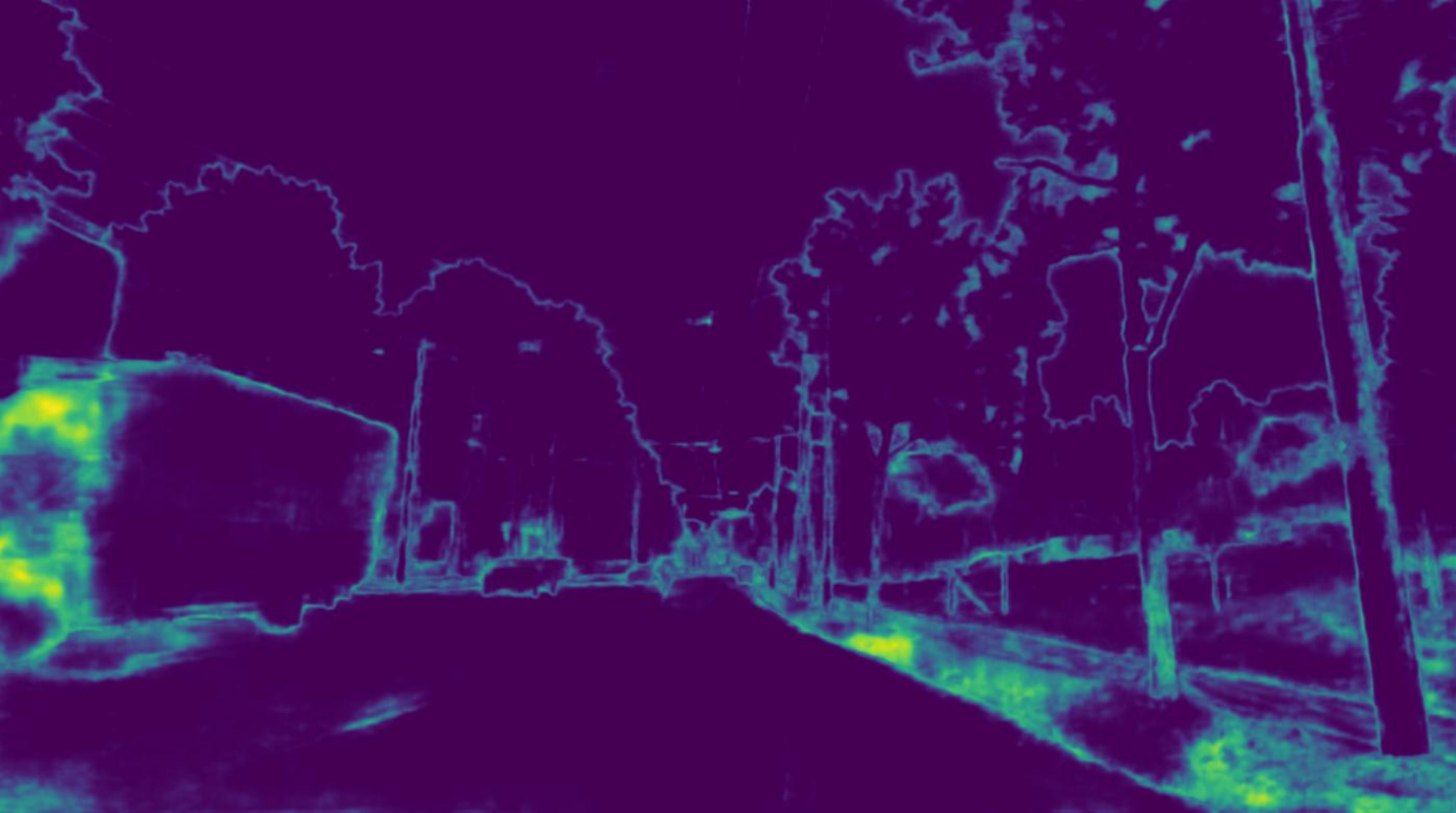}
    \end{subfigure}

    \begin{subfigure}{0.025\textwidth}
        \rotatebox[origin=c]{90}{U-CE$_{\alpha=10}$}\quad
        \phantom{\includegraphics[width=\textwidth,valign=c]{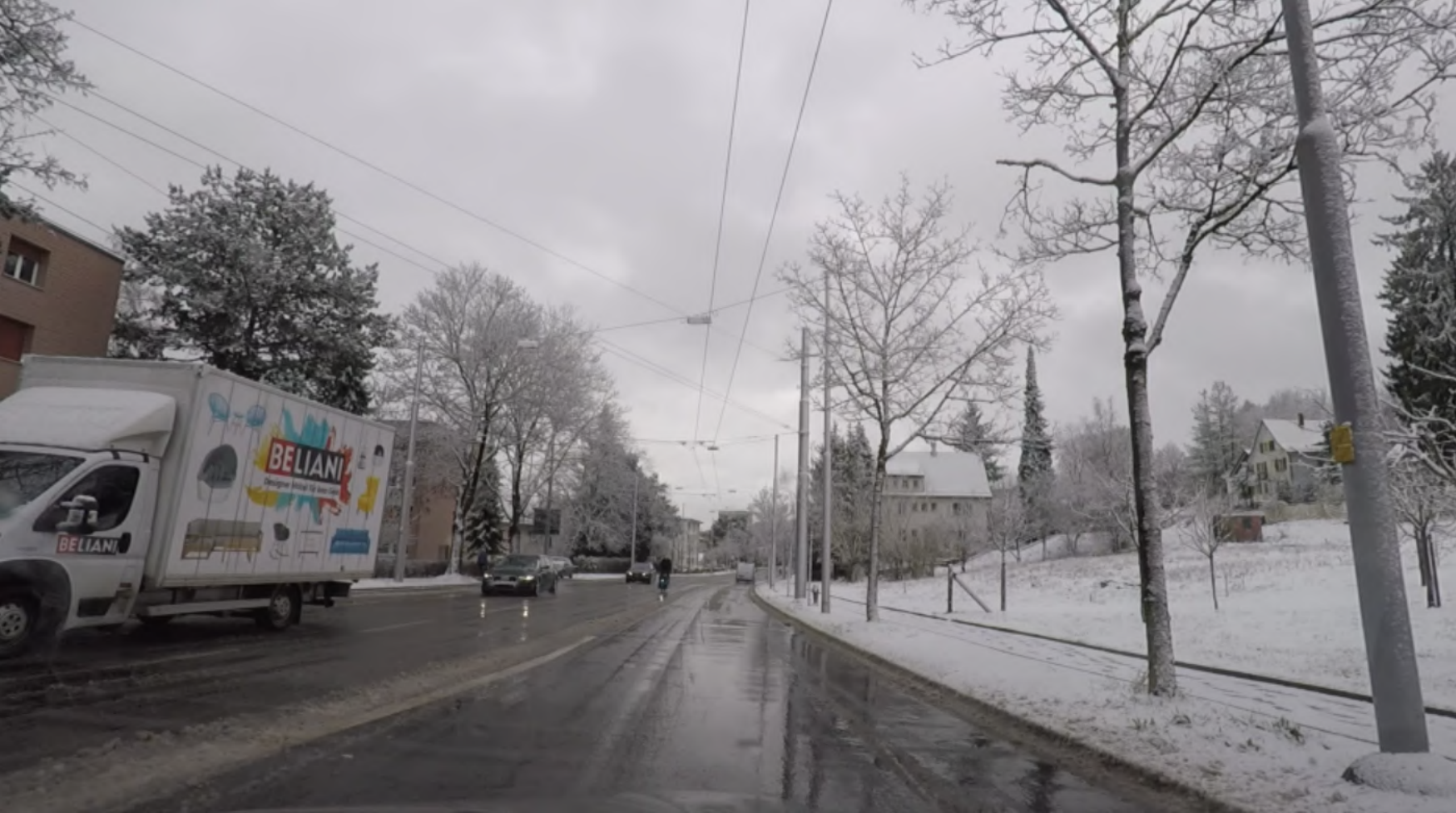}}
        \phantom{\includegraphics[width=0.1\textwidth,valign=c]{iccv2023AuthorKit/figures/acdc_uce_10_image-min.pdf}}
    \end{subfigure}\hfill
    \begin{subfigure}{0.19\textwidth}
        \includegraphics[width=\textwidth,valign=c]{iccv2023AuthorKit/figures/acdc_uce_10_image-min.pdf}
        \caption{Image}
    \end{subfigure}\hfill
    \begin{subfigure}{0.19\textwidth}
        \includegraphics[width=\textwidth,valign=c]{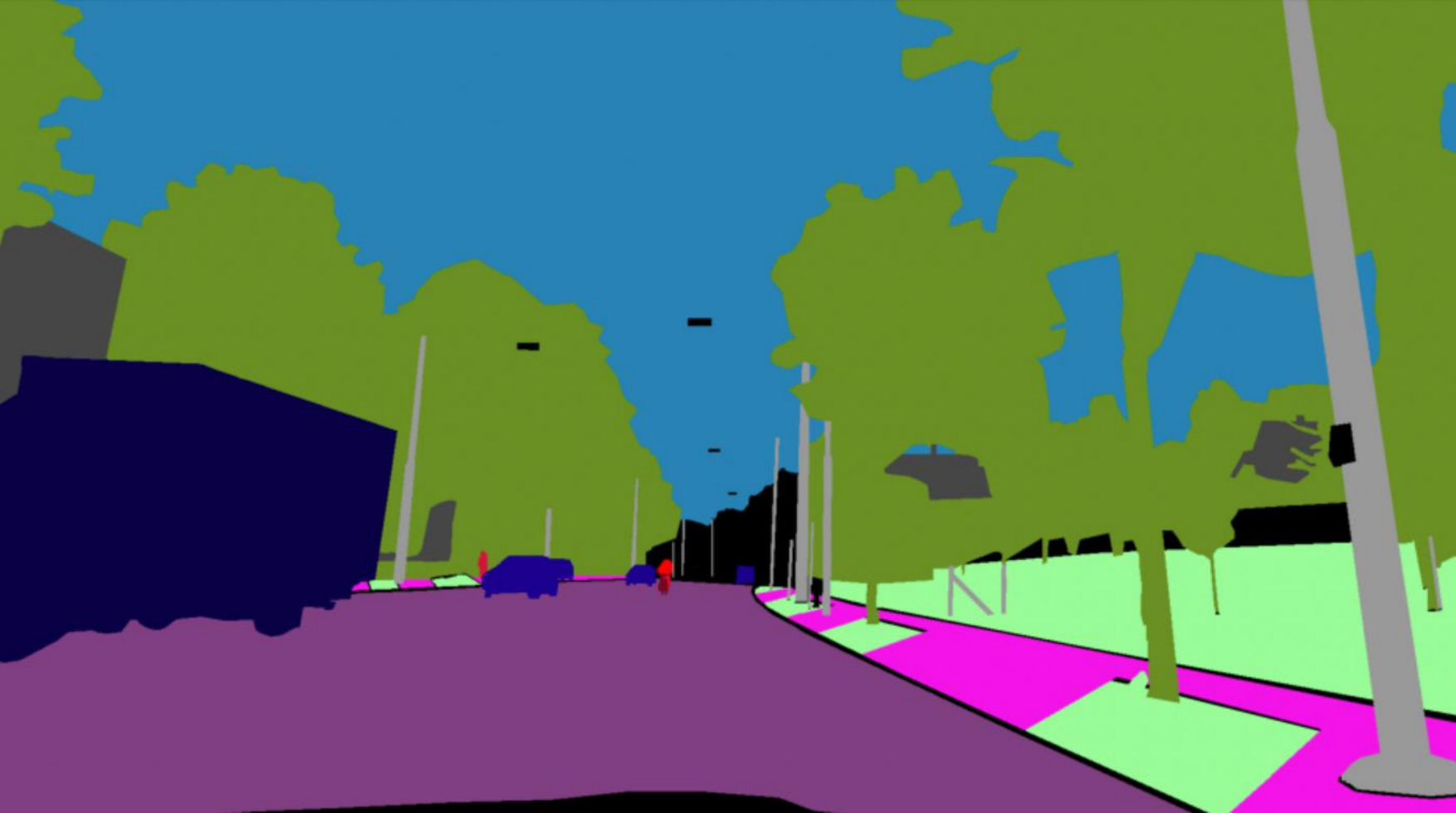}
        \caption{Ground Truth Label}
    \end{subfigure}\hfill
    \begin{subfigure}{0.19\textwidth}
        \includegraphics[width=\textwidth,valign=c]{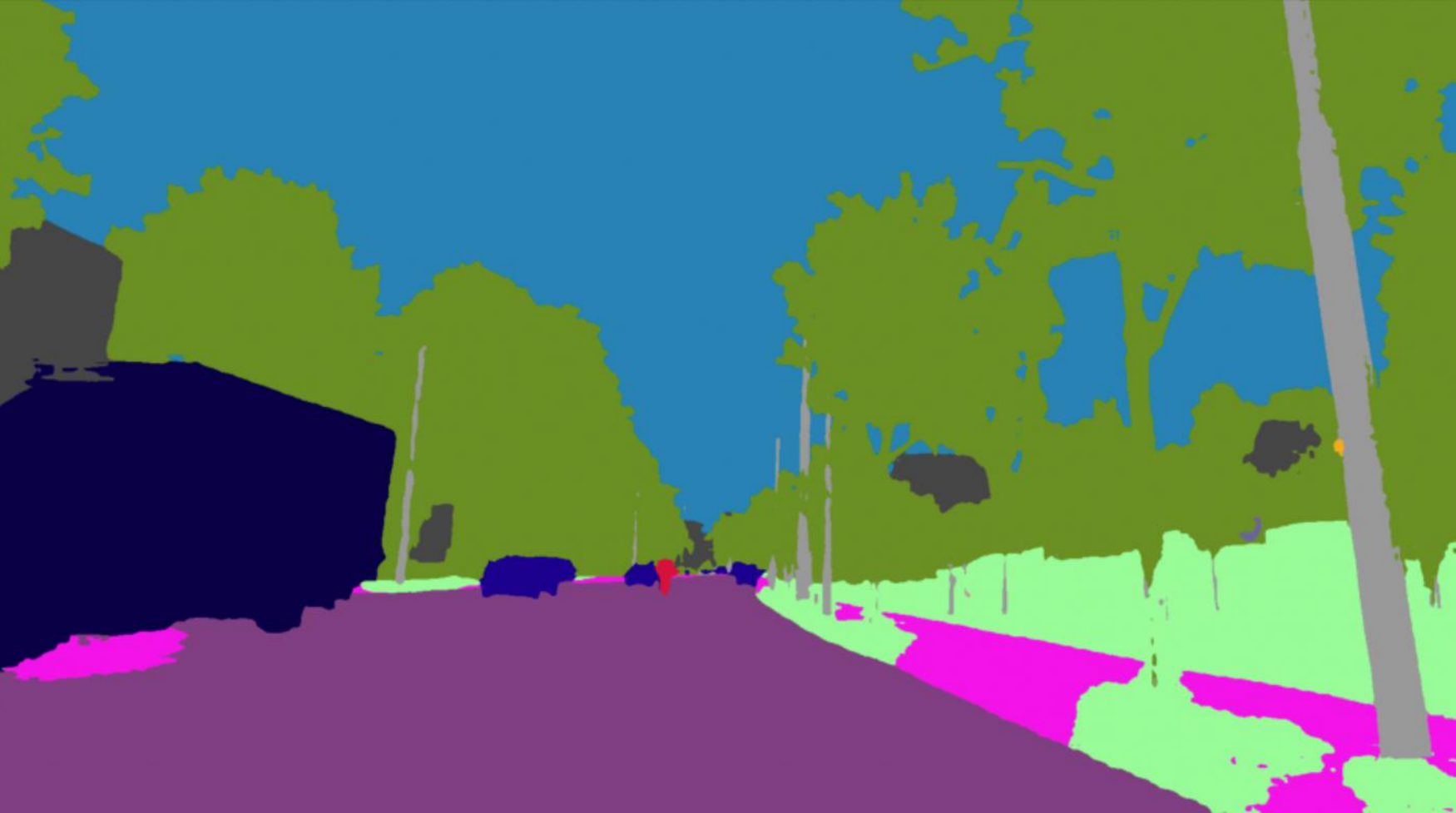}
        \caption{Segmentation Prediction}
    \end{subfigure}\hfill
    \begin{subfigure}{0.19\textwidth}
        \includegraphics[width=\textwidth,valign=c]{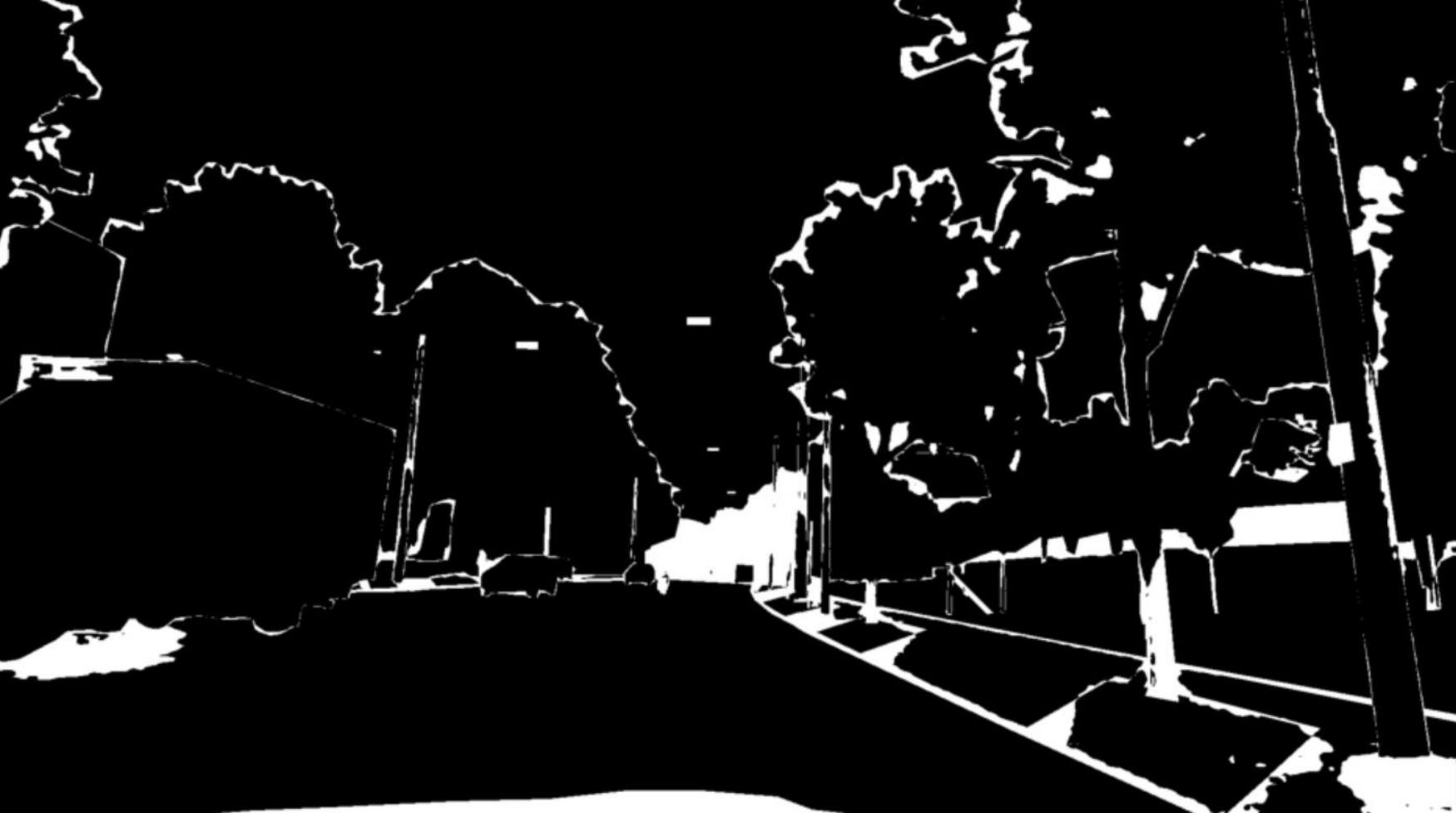}
        \caption{Binary Accuracy Map}
    \end{subfigure}\hfill
    \begin{subfigure}{0.19\textwidth}
        \includegraphics[width=\textwidth,valign=c]{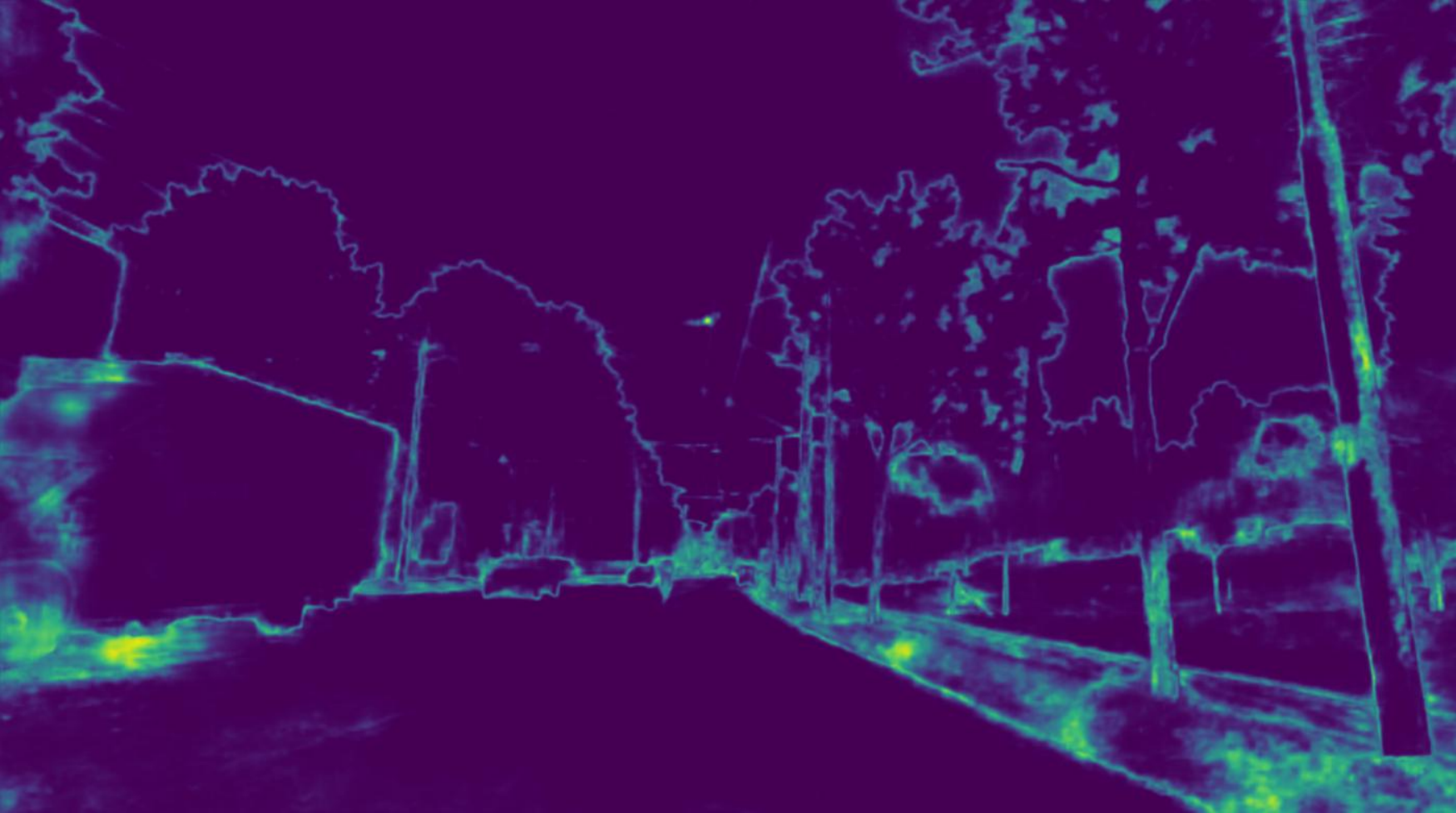}
        \caption{Predictive Uncertainty}
    \end{subfigure}
    
    \caption{Example images from the Cityscapes and ACDC validation set (a), corresponding ground truth labels (b), the model's segmentation predictions (c), a binary accuracy map (d), and the predictive uncertainty (e). White pixels in the binary accuracy map are either incorrect predictions or void classes, which appear black in the ground truth label. For the uncertainty prediction, brighter pixels represent higher predictive uncertainties. The first three rows depict results from models with a ResNet-18 backbone and dropout ratio of $20\%$, trained for 200 epochs on Cityscapes \cite{cordts2016CityscapesDataset}. The last three rows show examples from models using a ResNet-101 backbone and a dropout ratio of $20\%$, trained for 500 epochs on the ACDC dataset \cite{SDV21}.}
\label{fig: qualitative_results}
\end{figure*}

Tables \ref{tab:cityscapes} and \ref{tab:acdc} outline a quantitative comparison between regular CE and our proposed U-CE loss using two different $\alpha$ values for various dropout ratios and training lengths on the Cityscapes \cite{cordts2016CityscapesDataset} and ACDC \cite{SDV21} datasets. Remarkably, U-CE$_{\alpha=10}$ achieves the highest mIoU across all dropout ratios, even outperforming the baseline models that do not use dropout in most cases. Notably, U-CE$_{\alpha=10}$ achieves a maximum improvement of up to $9.3\%$ over regular CE when training on ACDC \cite{SDV21} for 200 epochs using a ResNet-18 with a dropout ratio of $40\%$. On average, U-CE$_{\alpha=10}$ outperforms CE by $2.0\%$ on Cityscapes \cite{cordts2016CityscapesDataset} and by $4.6\%$ on ACDC \cite{SDV21}. Interestingly, U-CE$_{\alpha=1}$ also matches or improves upon regular CE training in most cases. On average, U-CE$_{\alpha=1}$ outperforms CE by $0.3\%$ on Cityscapes and by $1.3\%$ on ACDC. 

Table \ref{tab:city_detailed} provides additional information on the ECE and mUnc for CE and U-CE using a dropout ratio of $20\%$. In comparison to regular CE and U-CE$_{\alpha=1}$, which exhibit similar results, U-CE$_{\alpha=10}$ not only improves segmentation performance but also yields slightly better calibrated networks, as measured by the ECE. Moreover, the mUnc is also slightly lower for U-CE$_{\alpha=10}$.

Overall, Tables \ref{tab:cityscapes}, \ref{tab:acdc} and \ref{tab:city_detailed} provide strong evidence for the effectiveness of leveraging predictive uncertainties in the training process.

\subsection{Qualitative Evaluation}
In addition to the quantitative evaluation, we also provide qualitative examples in Figure \ref{fig: qualitative_results} showing the original input image, the corresponding ground truth label, the model's segmentation prediction, a binary accuracy map, and the student's predictive uncertainty. The first three rows depict results from models with a ResNet-18 backbone and a dropout ratio of $20\%$, trained for 200 epochs with CE, U-CE$_{\alpha=1}$, U-CE$_{\alpha=10}$ on Cityscapes \cite{cordts2016CityscapesDataset}. The last three rows show examples from models using a ResNet-101 backbone and a dropout ratio of $20\%$, trained for 500 epochs on the ACDC dataset \cite{SDV21}. The binary accuracy map visualizes incorrectly predicted pixels and void classes in white, and correctly predicted pixels in black.

Generally, for large areas and well-represented classes like road, building, sky, and car, all models perform exceptionally well with minimal errors. Furthermore, there is a strong correlation between the binary accuracy map and the predictive uncertainty, indicating that all models provide meaningful uncertainties.

Nonetheless, there are nuanced differences between the models. For example, in the first two rows of Figure \ref{fig: qualitative_results}, which represent models trained CE and U-CE$_{\alpha=1}$, there are noticeable misclassifications on top of the human standing in front of the truck. Naturally, this area is also accompanied with high uncertainties. In contrast, the model trained with U-CE$_{\alpha=10}$ exhibits significantly fewer difficulties, resulting in a better segmentation prediction and lower uncertainties.

A similar situation is observable in the last three rows, showing examples from the more challenging ACDC dataset \cite{SDV21}. Here, the model trained with regular CE struggles to correctly segment the truck on the left as well as differentiate between the sidewalk and the terrain on the right side of the image. The model trained with U-CE$_{\alpha=1}$ does slightly better in these areas, but is equally uncertain. Only the model trained with U-CE$_{\alpha=10}$ successfully classifies the truck and differentiates between the sidewalk and the terrain decently. Consequently, the predictive uncertainty is also lower in these areas. 

In summary, the qualitative findings presented in Figure \ref{fig: qualitative_results} concur with our quantitative evaluation, manifesting the efficacy of U-CE across different datasets and architectures. 

\subsection{Ablation Studies}\label{sec:abl}
\begin{table}[t!]
\begin{center}
\begin{adjustbox}{width=\linewidth}
\setlength\extrarowheight{1mm}
\begin{tabular}{l|ccccccccc}
                                $\alpha$ & 1    & 2    & 4    & 6    & 8    & 10   & 12   & 14   & 16   \\ \hline
RN18 (Cityscapes)  & 69.5 & 70.0 & 70.7 & 71.2 & 71.5 & \textbf{71.8} & 71.0 & 47.0 & 70.9 \\
RN101 (Cityscapes) & 74.8 & 75.2 & 75.6 & 76.1 & 76.4 & \textbf{76.6} & 76.3 & 75.8 & 72.6 \\
RN18 (ACDC)        & 56.1 & 56.9 & 57.6 & 58.8 & 58.8 & \textbf{60.5} & 60.3 & 60.1 & 37.5 \\
RN101 (ACDC)       & 65.0 & 65.0 & 65.7 & 65.5 & 66.0 & 65.8 & \textbf{66.7} & 64.5 & 19.9    
\end{tabular}
\end{adjustbox}
\end{center}
\caption{Ablation study on the impact of $\alpha$. The provided numbers represent the mIoU $\uparrow$. Best respective results are marked in \textbf{bold}.}
\label{tab:abl_exp}
\end{table}

\begin{table}[t!]
\begin{center}
\begin{adjustbox}{width=\linewidth}
\setlength\extrarowheight{1mm}
\begin{tabular}{l|cccccc}
                                $\beta$ & 0 & 2   & 6   & 10  & 14   & 18 \\ \hline
CE                  & 69.0 (1:49) & - & - & - & - & - \\                            
U-CE$_{\alpha=10}$  & -           & 71.1 (1:52) & 71.6 (2:01) & 71.6 (2:27) & 71.6 (2:53) & 71.7 (3:17) \\
\end{tabular}
\end{adjustbox}
\end{center}
\caption{Ablation study on the number of segmentation samples $\beta$. In addition to the mIoU $\uparrow$, we provide the training time in hours:minutes $\downarrow$ in paranthesis.}
\label{tab:abl_samples}
\end{table}

\begin{table}[t!]
\begin{center}
\begin{adjustbox}{width=\linewidth}
\setlength\extrarowheight{1mm}
\setlength{\tabcolsep}{15pt}
\begin{tabular}{l|ccc}
     & Random Flipping & Random Scaling & mIoU $\uparrow$ \\ \hline
CE   & $\times$                & $\times$                         & 66.1 \\
     & \checkmark                & $\times$                & 67.0 \\
     & $\times$                & \checkmark                & 68.6 \\
     & \checkmark                & \checkmark       & 69.0 \\ \hline
U-CE$_{\alpha=1}$   & $\times$                & $\times$          & 65.8 \\
     & \checkmark                & $\times$                & 67.8 \\
     & $\times$                & \checkmark                & 69.1 \\
     & \checkmark                & \checkmark       & 69.5 \\ \hline
U-CE$_{\alpha=10}$   & $\times$                & $\times$         & 69.6 \\
     & \checkmark                & $\times$                & 70.1 \\
     & $\times$                & \checkmark                & 71.8 \\
     & \checkmark                & \checkmark       & 71.8 \\
\end{tabular}
\end{adjustbox}
\end{center}
\caption{Ablation study on the impact of various data augmentations strategies. We use random cropping with a crop size of $768 \times 768$ pixels as a baseline for all strategies.}
\label{tab:abl_aug}
\end{table}

\begin{table}[t!]
\begin{center}
\begin{adjustbox}{width=\linewidth}
\setlength\extrarowheight{1mm}
\setlength{\tabcolsep}{15pt}
\begin{tabular}{l|ccccc}
                                 $lr_{base}$ & $10^{-1}$   & $10^{-2}$  & $10^{-3}$ & $10^{-4}$ & $10^{-5}$ \\ \hline
CE                 & 50.5 & 69.0 & 55.9 & 35.6 & 18.9 \\
U-CE$_{\alpha=1}$ & \textbf{56.0} & 69.5 & 57.6 & 36.9 & 19.3 \\
U-CE$_{\alpha=10}$ & 2.0 & \textbf{71.8} & \textbf{65.0} & \textbf{47.6} & \textbf{25.3} \\
\end{tabular}
\end{adjustbox}
\end{center}
\caption{Ablation study on the base learning rate $lr_{base}$. The provided numbers represent the mIoU $\uparrow$. Best results are marked in \textbf{bold}.}
\label{tab:abl_lr}
\end{table}

In addition to the quantitative and qualitative evaluation, we also present multiple ablation studies. Unless otherwise noted, we confined all of the ablation studies to models that use a ResNet-18 as the backbone, have a dropout ratio of 20\%, and were trained for 200 epochs.

\textbf{Impact of $\alpha$.} The most influential hyperparameter of U-CE is $\alpha$ as it exponentially controls the weighting of the CE loss. Table \ref{tab:abl_exp} demonstrates the impact of different $\alpha$ values on the mIoU for both backbones, ResNet-18 (RN18) and ResNet-101 (RN101), on both Cityscapes and ACDC. Evidently, the segmentation performance consistently improves as $\alpha$ increases until it reaches ten, which stands as the best value in three out of four cases across the two datasets and architectures. Thus, using ten as the default value for $\alpha$ seems to be a fair estimation to achieve the best results, not only for the mentioned cases but potentially for other applications as well. Further increasing $\alpha$ leads to a degradation in mIoU. Additionally, training becomes more unstable as models overly focus on uncertain pixels, resulting in some models failing to converge properly. Nonetheless, U-CE exhibits robustness against changes in $\alpha$, offering a wide range of valid hyperparameters that lead to improved segmentation results compared to regular CE training.

\textbf{Impact of $\beta$.} Table \ref{tab:abl_samples} exhibits another ablation study on the number of segmentation samples $\beta$. Interestingly, there is no clear benefit of sampling more often than six times, especially with regard to the training time. As indicated by the training times, U-CE$_{\beta=6}$ increases the necessary training time by approximately $10\%$, whereas U-CE$_{\beta=10}$ extends it by roughly $35\%$. For comparison, Gal and Ghahramani \cite{gal2016DropoutBayesian} recommend sampling ten times to get a reasonable estimation of the predictive mean and uncertainty. 

\textbf{Impact of Data Augmentations.} The impact of various data augmentation strategies on CE and U-CE is demonstrated in Table \ref{tab:abl_aug}. The results show that incorporating additional data augmentations on top of the baseline strategy of random cropping with a crop size of $768 \times 768$ pixels improves the mIoU across the board. More importantly, this ablation study confirms that U-CE consistently outperforms CE across different data augmentation strategies, indicating its effectiveness in improving segmentation performance.

\textbf{Impact of $lr_{base}$.} Table \ref{tab:abl_lr} shows the ablation study on the base learning rate $lr_{base}$. The most notable comparison is between regular CE and U-CE$_{\alpha=1}$, which demonstrates that U-CE is not limited to specific learning rates. U-CE$_{\alpha=1}$ consistently outperforms regular CE for all examined base learning rates, despite increasing the training loss by approximately $9\%$ as indicated by the mUnc in Table \ref{tab:city_detailed}.  Moreover, U-CE$_{\alpha=10}$ exceeds the results of CE and U-CE$_{\alpha=1}$ for all base learning rates except $10^{-1}$, which caused divergence. Overall, this ablation study confirms the value of leveraging predictive uncertainties during training, irrespective of the learning rate, which is arguably the single most important hyperparameter in deep learning \cite{bengio2012practical}.

\section{Discussion}\label{sec: discussion}
In contrast to previous approaches, U-CE fully leverages predictive uncertainties obtained by Monte Carlo Dropout during training. As a result, we manage to train models that not only improve their segmentation performance but are also naturally capable of predicting meaningful uncertainties after training as well. 

While U-CE appears to have no apparent shortcomings, except for a minor increase in training time, we acknowledge the need for a transparent discussion about its potential limitations. Our aim is to effectively guide future work in pushing the boundaries of state-of-the-art techniques, especially in safety-critical applications like autonomous driving.

\textbf{Limitations.} One limitation of U-CE arises in the absence of densely annotated ground truth labels. If most pixels are either labeled as background or designated to be ignored while training, U-CE will likely offer next to no benefit, except for a higher loss around object boundaries. Additionally, U-CE may not contribute to improved segmentation performance if the network is already overfitting the training data. Having said that, the impact of U-CE on generalization needs further examination. 

\textbf{Future Work.} With regards to future work, we have multiple suggestions that might be worth investigating. Potentially, the results of U-CE could be further improved if the quality of the uncertainty estimates would be better. 
Therefore, it would be interesting to integrate Deep Ensembles \cite{lakshminarayanan2017SimpleScalable}, the state-of-the-art uncertainty quantification method \cite{ovadia2019DatasetShift, wursthorn2022, gustafsson2020evaluating}, with U-CE, which we could not realize because of computational restrains. On a similar note, it could be worth employing warmup epochs, which we omitted to refrain from introducing another hyperparameter. Additionally, we would like to see $\alpha$ removed from U-CE by incorporating statistical hypothesis testing. This would be beneficial in two ways: Firstly, it would remove the most influential hyperparameter of U-CE. Secondly, and maybe more importantly, it would leverage all of the available uncertainties and not just the predictive uncertainty. Finally, we encourage other researchers to incorporate U-CE into state-of-the-art semantic segmentation approaches and to explore its usefulness in other computer vision tasks that rely on pixel-wise predictions, such as depth estimation.

Overall, we believe that U-CE presents a promising paradigm in semantic segmentation by dynamically leveraging uncertainties to create more robust and reliable models. Despite a minor increase in training time and room for further improvement, we see no reason not to employ U-CE in comparison to regular CE.

\section{Conclusion}
In this paper, we introduced U-CE, a novel uncertainty-aware cross-entropy loss for semantic segmentation. U-CE incorporates predictive uncertainties, based on Monte Carlo Dropout, into the training process through pixel-wise weighting of the regular cross-entropy loss. As a result, we manage to train models that are naturally capable of predicting meaningful uncertainties after training while simultaneously improving their segmentation performance. Through extensive experimentation on the Cityscapes and ACDC datasets using ResNet-18 and ResNet-101 architectures, we demonstrated the superiority of U-CE over regular cross-entropy training.

We hope that U-CE and our thorough discussion of potential limitations and future work contribute to the development of more robust and trustworthy segmentation models, ultimately advancing the state-of-the-art in safety-critical applications and beyond.

\section*{Acknowledgment}
The authors acknowledge support by the state of Baden-Württemberg through bwHPC.

This work is supported by the Helmholtz Association Initiative and Networking Fund on the HAICORE@KIT partition.

{\small
\bibliographystyle{ieee_fullname}
\bibliography{egbib}
}

\end{document}